\RequirePackage{fix-cm}
\documentclass[smallextended]{svjour3}       
\smartqed  
\usepackage{graphicx}
\usepackage{booktabs}
\usepackage{pifont}
\usepackage[justification=centering]{caption}
\usepackage{svg}
\newcommand*\rot{\rotatebox{90}}

\usepackage{algorithm}
\usepackage{algpseudocode}
\usepackage[backend=biber,style=apa,sorting=nyt]{biblatex}
\usepackage{mathtools}
\usepackage{longtable}
\DeclareUnicodeCharacter{0301}{\'{e}}
\DeclareUnicodeCharacter{200B}{{\hskip 0pt}}
\pdfoutput=1

\begin{document}

\title{DISCO PAL
}
%

\subtitle{Diachronic Spanish Sonnet Corpus with Psychological and Affective Labels}


\author{Alberto Barbado         \and
        Víctor Fresno \and Ángeles Manjarrés Riesco \and Salvador Ros
}


\institute{Alberto Barbado \at
              Telefónica, 28050 Madrid, Spain \\
              Universidad Nacional de Educación a Distancia (UNED) \\
              \email{alberto.barbadogonzalez@telefonica.com}           
           \and
           Víctor Fresno \at
              Universidad Nacional de Educación a Distancia (UNED) \\
              \email {vfresno@lsi.uned.es}
            \and
           Ángeles Manjarrés Riesco \at
              Universidad Nacional de Educación a Distancia (UNED) \\
              \email {amanja@dia.uned.es}  
            \and
           Salvador Ros \at
              Universidad Nacional de Educación a Distancia (UNED) \\
              \email {sros@dia.uned.es}
}

\date{Received: date / Accepted: date}

\maketitle

\begin{abstract}
Nowadays, there are many applications of text mining over corpora from different languages. However, most of them are based on texts in prose, lacking applications that work with poetry texts.

An example of an application of text mining in poetry is the usage of features derived from their individual words in order to capture the lexical, sublexical and interlexical meaning, and infer the General Affective Meaning (GAM) of the text. However, even though this proposal has been proved as useful for poetry in some languages, there is a lack of studies for both Spanish poetry and for highly-structured poetic compositions such as sonnets.

This article presents a study over an annotated corpus of Spanish sonnets, in order to analyse if it is possible to build features from their individual words for predicting their GAM. The purpose of this is to model sonnets at an affective level. 

The article also analyses the relationship between the GAM of the sonnets and the content itself. For this, we consider the content from a psychological perspective, identifying with tags when a sonnet is related to a specific term. Then, we study how GAM changes according to each of those psychological terms. 

The corpus used contains 274 Spanish sonnets from authors of different centuries, from 15th to 19th.  This corpus was annotated by different domain experts. The experts annotated the poems with affective and lexico-semantic features, as well as with domain concepts that belong to psychology. Thanks to this, the corpus of sonnets can be used in different applications, such as poetry recommender systems, personality text mining studies of the authors, or the usage of poetry for therapeutic purposes. 
\end{abstract}

\section{Introduction}
\label{intro}
\noindent Text mining techniques aim to extract insights from a text and discover patterns within it using different kinds of information from that text. As an example, the information contained in a text could be related to its syntactical structure, to its semantical meaning, or it can even consider information sources such as its affective value (e.g. if a text inspires a certain emotion when read). 
This is the base for many pieces of research, such as sentiment analysis. Sentiment analysis, also called "opinion mining", is the field of study that analyses people's opinions, feelings, assessments, attitudes, and emotions towards entities such as products, services, organizations, individuals, problems, events, topics and their attributes \parencite{liu2012sentiment}.

Thus, this is applicable to the field of text mining, where these feelings can be inferred using input information derived from the text itself. This is the case of the inference of the General Affective Meaning (GAM) \parencite{aryani2016measuring} of the text. In a text, GAM can be obtained from its semantic information, the affective information of the individual words which compose it, the type of text used and its syntactic characteristics… That initial information generates features that represent a text, and those features serve as an input for a function that yields an output corresponding to GAM tags. An example of such features are valence or arousal \parencite{russell2003core, tsur1992makes, watson1985toward, wundt1874grundzuge}.

Regarding those functions, they can use prior information about the GAM tags (supervised) or not (unsupervised). An example for the first case is the usage of Supervised Machine Learning (ML) algorithms. Here, we need to know the GAM of some texts in order to train the supervised ML model with the input features so as to be able to obtain the GAM for the texts where it is not known.

However, those approaches are not applicable for the unsupervised scenario where there are no GAM tags available.

GAM are not the only type of variables that can be used to model the global meaning of a text. Other variables may be considered, including words related to the semantic meaning of the text, such as the relation between definitions and their associated words in a dictionary \parencite{noraset2017definition}.

All the different kinds of information contained within a text (semantic, syntactic, affective…) depend on the type of text considered. Because of this, the approach will be different depending on whether the text is, for example, prose or verse. It will also depend on the language used in the text.

That said, there are not many corpora available to perform data mining tasks on poetry texts, and much less for the Spanish language. There are even fewer options related to GAM and poetry. 
It is true that there are available corpora for Spanish poetry, such as the corpus DISCO \parencite{ruiz2018disco}, but the annotations included within it do not provide information that can be used directly for text modelling tasks, such as obtaining the GAM. The reason is that DISCO only includes metadata about authors, sonnet scansion, rhyme-scheme and enjambment. 

This is the reason why the present research increases the available copora for text mining tasks with Spanish poetry by presenting a corpus of Spanish sonnets from different time periods annotated with both affective and lexico-semantic labels in order to contribute to the research of text mining in both areas. The article will present DISCO PAL, Diachronic Spanish Sonnet Corpus with Psychological and Affective Labels (together with this paper), a corpus annotated by POSTDATA\footnote{Poetry Standardization and Linked Open Data, Ref. ERC-2015-STG-679528 proyect Starting Grant from European Research Council within the horizon H2020.} experts in both literature and digital humanities. POSTDATA project aims to make "poetry available online as machine-readable data will open a great world of possibilities of linking, indexing and extracting new information.".
This corpus includes binary labels for a group of concepts depending on whether that concept appears within the text or not. The concepts used all belong to the psychological domain.

Overall, the main contributions of this article are to:
\begin{itemize}
    \item Define a methodology for unsupervised GAM modelling of a corpus of Spanish sonnets, based on previous works of GAM modelling for poetry in other languages. The proposal uses as input data sources public lexicons with the affective meaning of individual words in Spanish in order to build affective and semantic features that infer the GAM for the whole text. 
    \item Validate the unsupervised GAM proposal by using an annotated corpus of Spanish sonnets (DISCO PAL) by different domain experts. This corpus contains annotations for the same features generated by the GAM modelling. The annotations values depend on the intensity of that variable within each sonnet.
    \item Analyse how the content influence the GAM generated. For this, the experts also annotated values for labels of psychological concepts that are expressed through that sonnet. 
    \item Provide the DISCO PAL corpus for future research, highlighting possible ways to use it for data mining on poetry, mainly through the affective and semantic modelling of texts.
\end{itemize}

The structure of this article is as follows: after the Introduction presented in this first Section, the second Section summarizes the state of the art (SOTA) for the areas relevant for this article. First, the SOTA related to data mining of affective information on poems. Then, the SOTA related to affective modelling of Spanish language by using public lexicons for modelling individual words.
After that, the third Section presents the DISCO PAL corpus annotated by POSTDATA experts in digital humanities, analysing the agreement between the annotators and the reliability of the corpus. It also follows the research applied on poetry in other languages in order to build features based on the affective and lexico-semantic values of individual words (using public lexicons for affective and lexico-semantic word modelling in Spanish). Then, we see if those features generated capture the GAM of the sonnets by checking the values inferred against the ones annotated by the POSTDATA experts.
The last Section mentions the potential lines of research that could be carried out thanks to this corpus. It also includes a summary with the conclusions of this article.

\section{Related Work}
\label{sec:1}
This Section presents a brief review of the related work. As we mentioned in the Introduction, a text contains information related to different areas such as semantics, syntactic or affective states. This is applicable to all kinds of texts, including poetic ones. Since this article provides a research related to the affective modelling of Spanish poetry, the main area covered in this Section is related to the data mining process of affective information in poetry. We complement this with a subsection describing some public lexicons for the affective modelling of individual Spanish words. 

\subsection{Data mining of affective information in poetry}
As previously indicated, texts in general, and particularly poetic ones, contain affective information that can be extracted using different techniques. An example of this is aggregating the affective values of the individual words present in the text. It is important to quantify this affective contribution of poetic texts in order to work with them computationally. Thus, the task consists of detecting which elements of the text are especially relevant in order to calculate through them the affective contribution of the whole poem. The articles shown below analyse different ways of extracting and quantifying affective aspects from poetic texts.

In order to model the GAM of a poetry text, that poem needs to be expressed through a set of relevant features that are linked to GAM using a relationship that is expressed with a mathematical function. From here, there are two possibilities. First, if there is information about the GAM value within some poems, the objective function may consist in generalizing the relationship between those values and the features extracted from the poem. This can be used later on to infer the GAM in poems where there is no prior information about it (and only the features extracted are available). This scenario is approached in \parencite{sreeja2018emotion} with the usage of supervised ML models. Here, the authors provide a corpus of 736 English poems annotated with 9 affective labels (love, anger, hate, sadness, joy, surprise), and use it to train an ensemble of supervised ML models. They begin extracting a set of relevant features from the poems related to semantic, linguistic and orthographic aspects, as well as some statistical features (term frequency and inverse document frequency). They also use poetic features extracted with rule-based methods which include information related to simile and metaphors. Those features are used together with the annotated affective labels in order to train ML models that can predict the GAM value for new sonnets. It is also worth mentioning how the authors state that this article is "the first attempt to identify emotions from English poems".

A similar approach was considered before for Arabic poetry in \parencite{alsharif2013emotion}. Here, the authors built a corpus of Arabic poems annotating them with different emotions. Then, they extracted a set of relevant features from the poems based on the occurrences of different words within them (unigrams). With these feature vectors, they trained different supervised ML models (Support Vector Machines, Naïve Bayes, Voting Features Intervals and Hyperpipes) to predict the emotion values.

In \parencite{jacobs2017s}, the authors first use a Quantitative Narrative Analysis tool and 11 questions to find relevant features that appear and model 154 Shakespeare's sonnets. These features include affective variables (such as the valence or the mood potential). Then, they use a ML model trained with these features to classify the models into two categories, "young man" poems or "dark lady" ones.

There are some available corpora of sonnets that include affective annotations by experts that can be used for tasks like the ones mentioned before. In \parencite{sreeja2019perc}, the authors built a benchmark corpus for poetry named PERC (Poem Emotion Recognition Corpus). This corpus includes poems from Indian authors in English. For those poems, they include 9 emotions that are annotated in the corpus by several experts (Love,
Sadness, Joy, Fear, Hate, Courage, Anger, Surprise and Peace). The corpus provided by the authors includes 1850 poems from 10 authors from 1850 to the present day.
Similarly, in \parencite{haider2020po}, the authors provide an annotated corpus that includes 158 German poems along with 64 poems in English. The annotations include 9 affective features (Vitality, Uneasiness, Suspense, Sadness, Nostalgia, Humor, Beauty/Joy, Awe/Sublime and Annoyance).

Beyond these supervised proposals, other authors have tackled the problem in an unsupervised manner. In \parencite{barros2013automatic}, the authors obtain the GAM by counting how many instances of words such as \textit{fear} or \textit{joy} appear within a set of Quevedo’s poems. Therefore, no prior annotated values are used to infer the GAM of a poem. The final extracted GAM values are used to automatically annotate that corpus.

This last paper, in fact, deals with the GAM extraction from Spanish poems. However, there are no more corpora beyond this one to the best of our knowledge. In fact, more recent research studies of the topic for data mining with poetry, such as \parencite{kaur2017punjabi} or \parencite{sreeja2019perc}, only list that corpus of Quevedo’s poems annotated with sentiment labels according to the presence of certain words as Spanish corpora sources for data mining and GAM modelling.

Features can be obtained by modelling the whole text or by modelling the individual stanzas of the poem. For the case of affective features, they can be inferred using as input the individual affective values of the words that appear in the text as long as there is an available lexicon that contains those individual affective values, such as BAWL in \parencite{ullrich2017relation}. The authors perform data mining of affective content in poetic texts for German language. The authors explore how the features of a poetic text (at sub-lexical, lexical and inter-lexical level) influence the GAM that is perceived. Thus, their research serves as an example to see which affective features are relevant to a text based on how related they are to the GAM. To calculate those features they use the BAWL database for German words. This database contains affective values for individual words that belong to German, and they aggregate these individual values into a global value that models the GAM for the whole poem.

Their poem corpus consists of 57 poems from the German author H.M. Enzenberg. These poems are annotated by a group of readers with the following features:

\begin{enumerate}
  \item Rating on a scale of 7 for the valence, where -3 would be very negative, 0 neutral and 3 very positive.
  \item Rating on a scale of 5 for the arousal (level of excitement of the text of the poem, which goes from texts that inspire peacefulness or calmness, to others that seek to motivate or are more exciting), where 1 is very calm and 5 very exciting.
  \item Rating on a scale of 1 to 5 for the level of friendliness, where 1 indicates that the text is not friendly and 5 that it is very friendly.
  \item Rating on a scale of 1 to 5 for the level of sadness, where 1 would be that the text is not sad, and 5 that it is very sad.
  \item Rating from 1 to 5 for the level of malevolence.
  \item Rating from 1 to 5 indicating if they liked the poem a lot or not (5, a lot).
  \item Rating from 1 to 5 for the level of poeticity, where 5 would indicate that the poem is very poetic and 1 that it is not very poetic.
  \item Rating from 1 to 5 for the level of onomatopoeia (level that quantifies the use of this literary resource). 5 would indicate a very frequent usage.
\end{enumerate}

These annotations by users at a global level serve to analyse their correlation against different features derived from the individual words that appear within the text (not considering stopwords). The purpose of this study is to check if the features could be used for predicting the GAM of the poem.
As mentioned before, the features belong to three different levels: sub-lexical, lexical and inter-lexical. The lexical level captures the average valence and arousal values from the words present in the text, the inter-lexical level quantifies peaks, ranges and changes within the lexical affective content, and the sub-lexical level considers aspects such as the phonological information of the poems. 
All this defines 55 affective features (using the 3 levels described above). Approximately the 50 percent of the explained variance is reached using only the lexical features, and together with the inter-lexical ones, the explained variance reaches 75 percent. This indicates that the best predictors would be the ones related to these two levels, particularly the average of valence and the average of arousal derived from the individual words.

Of course, considering only these two features would indicate that the order of the words in the text is irrelevant for the affective impact, and that is not the case; the order matters, and experiencing crescendos or affective decrescendos is something fundamental, so the span of the level of excitation is another key aspect to take into account. Together with that, the article also considers how the valence and arousal levels evolve during the poem. This is important because, for example, poems are generally perceived as sadder when the valence of words is decreasing (more negative) and when the arousal at the end is lower, and poems are perceived as friendlier when the valence of the last words of the text is more positive. In this way it is important to consider the correlation coefficient between the vector of affectivity (arousal / valence) of the individual words with the vector of their positions in the text.

We find this article particularly relevant for our studies since it presents a thorough methodology for GAM extraction that concludes in good results. 

It is important to remember that poetry is a huge genre, where there are different types of styles, and that will influence the affective modelling and the GAM extracted. This is indicated in the work of \parencite{obermeier2013aesthetic}, where a study of the influence of poetry on affective states is presented thanks to certain aesthetic and emotional elements such as the metric of the poem and its rhyme. Thus, the starting hypothesis is that metrics and rhyme have an impact on aesthetic perception, emotional involvement and valence. This indicates that the GAM of a poem will vary depending on whether the poem's style includes metric and rhyme or not. 
To verify this, the authors analyse the influence of metrics and rhyme in the aesthetic and emotional perception of poetry, as well as their interaction with the lexicon, using the stanzas of the poems as references for the study. For that, they work with a group of 60 adults who listened to audios of German poems (100 poems from the 19th and 20th centuries). The poems had stanzas of 4 verses in which there were sets of poems with lexical differences (for instance, real words vs pseudowords; pseudowords were words modified that kept the vowels but changed some consonant, ensuring that they were still pronounceable). Poems also were divided depending on whether they had rhymes or not, or if they had accent or not.
With this, the users rated four metrics for the poems that they were listening to: liking (aesthetic appreciation), intensity (power of emotional response), perceived emotion (emotion that was expressed within the stanza) and felt emotion (emotion experienced by the users).

The results are as follows:
\begin{itemize}
  \item \textbf{Liking}: results had better aesthetic ratings for poetry with metric as well as for stanzas with rhyme compared to those without it.
  \item \textbf{Intensity rating}: for all kind of poems the results were better with the stanzas that contain real words and not pseudowords.
   \item \textbf{Perceived emotion}: influence of lexicon, metric and rhyme (especially the last two); best score for stanzas with pseudowords if they don't have metric versus those that do have it. This last difference does not appear for poems with only real words.
  \item \textbf{Felt emotion}: the main influence is the rhyme. There is also a triple interaction between lexical-metric-rhyme. When there is rhyme the emotion felt is stronger.
\end{itemize}

Thus, this means that metric and rhyme reinforce the perceived emotion of a poem, which is expressed through the GAM. This serves as a basis to consider sonnets as good candidates for our studies regarding GAM extraction, since they are structured poems with rhyme and metric. Due to this, we will focus our analyses not only on Spanish poetry, following some of the steps of \parencite{ullrich2017relation}, but particularly in sonnets, as they will always guarantee the metric structure that enhances the text GAM.

As a last comment, however, the literature indicates some caveats and difficulties regarding the affective modelling of poetry. This appears in \parencite{eastman2015making}, where the authors propose a solution for affective computing in relationship to poetry. This article addresses two relevant issues in this regard. On the one hand, it reminds us how poetry widely uses metaphors and figurative language (words open to many meanings and interpretations). This makes the extraction of affective information not always as obvious as simply assigning to each word a value contained in a repository and then composing all the individual values. Metaphors are also interpreted in a large part from the subjectivity of the reader and from their personal experience, so it is not trivial and immediate to incorporate all the possible information. 
On the other hand, it also mentions that the understanding of the words of a poetic text should not only be done based on the text itself; a poem by a given author can be understood in greater depth when compared with other poems by that author or with poems of other authors. 
Due to this, it is important to note that the comprehension of a text, and hence the context for the individual words, is best achieved if the words are understood not only within the context of a specific poem or a specific author but in a bigger context that includes poems from other authors. This is something important in any text comprehension task, but it is even more critical for poems where the language used is sometimes full of metaphors and other stylistic figures not so easily understood. The proper comprehension of the text is important for both the semantic modelling of the poem as well as for the affective one, which means that the GAM extraction will be influenced by the context considered.

These previous works show how GAM extraction for poetry is tackled both with supervised and unsupervised approaches, covering poems from many different languages. However, there are few studies regarding Spanish poetry, with no references to sonnets in particular. Just like there are works that both analyse and provide an annotated poetry corpus with GAM values for German, Arabic and English texts, there are no equivalent, to the best of our knowledge, for Spanish. Therefore, we find a research need regarding both GAM extraction process for Spanish poetry, as well as offering an annotated corpus for future researches. Due to this, we focus our analysis in Spanish poetry, using sonnets in particular because of their stable structure and the presence of meter and rhyme. We follow the steps of \parencite{ullrich2017relation}, since they reach good results in the GAM extraction process while also referring to an annotated corpus. We will extract the GAM for sonnets in an unsupervised manner, and check the quality of those GAM values comparing the results against their counterpart values annotated by different experts.

\subsection{Lexicons for affective word modelling in Spanish}
Just like BAWL, as mentioned in \parencite{ullrich2017relation}, is a lexicon used as source information for the affective modelling of individual words, there are similar lexicons for Spanish vocabulary. Some of these lexicons are described below.

In \parencite{ferre2017moved} 2267 words are written in Spanish (along with their English translation) with the following variables\footnote{All the variables have ranges from 1 (minimum) to 5 (maximum).}:

\begin{itemize}
  \item \textit{Spanish\textunderscore Word}: word in Spanish.
  \item \textit{English\textunderscore Translation}: English translation of that word.
  \item \textit{Hap\textunderscore Mean}: mean value associated with this affective state (happiness) for that word from the individual ratings given by the subjects.
  \item \textit{Hap\textunderscore SD}: standard deviation associated with this affective state (happiness)  for that word from the individual ratings given by the subjects.
  \item \textit{Ang\textunderscore Mean}: idem for this affective state (anger).
  \item \textit{Ang\textunderscore SD}: idem for this affective state (anger).
  \item \textit{Sad\textunderscore Mean}: idem for this affective state (sadness).
  \item \textit{Sad\textunderscore SD}: idem for this affective state (sadness).
  \item \textit{Fear\textunderscore Mean}: idem for this affective state (fear).
  \item \textit{Fear\textunderscore SD}: idem for this affective state (fear).
  \item \textit{Disg\textunderscore Mean}: idem for this affective state (disgust).
  \item \textit{Disg\textunderscore SD}: idem for this affective state (disgust).
  \item \textit{N}: number of subjects used in the sample.
\end{itemize}

In \parencite{guasch2016spanish} 1400 words are written in Spanish with the following variables:
\begin{itemize}
    \item \textit{ID}: auto incremental field.
    \item \textit{Word}: word in Spanish.
    \item \textit{English Trans.}: English translation of the words. 
    \item \textit{POS}: Part of Speech (POS) tag for that word.
    \item \textit{VAL\textunderscore M}: mean value of the valence for that word from the individual ratings given by the subjects.
    \item \textit{VAL\textunderscore SD}: standard deviation of the valence for that word from the individual ratings given by the subjects
    \item \textit{VAL\textunderscore N}: number of subjects used to obtain valence values.
    \item \textit{ARO\textunderscore M}: idem for excitation level.
    \item \textit{ARO\textunderscore SD}: idem for excitation level.
    \item \textit{ARO\textunderscore N}: idem for excitation level.
    \item \textit{CON\textunderscore N}: idem for concreteness.
    \item \textit{CON\textunderscore SD}: idem for concreteness.
    \item \textit{CON\textunderscore N}: idem for concreteness.
    \item \textit{IMA\textunderscore M}: idem for imageability.
    \item \textit{IMA\textunderscore SD}: idem for imageability.
    \item \textit{IMA\textunderscore N}: idem for imageability.
    \item \textit{AVA\textunderscore M}: idem for context availability.
    \item \textit{AVA\textunderscore SD}: idem for context availability.
    \item \textit{AVA\textunderscore N}: idem for context availability.
    \item \textit{FAM\textunderscore M}: idem for familiarity.
    \item \textit{FAM\textunderscore SD}: idem for familiarity.
    \item \textit{FAM\textunderscore N}: idem for familiarity.
\end{itemize}

Regarding the concepts used, Concreteness is defined as the degree of specificity of the word, with 1 representing when the word is very abstract and 7 when it is very concrete. Words like ‘object’ are more abstract than others like ‘table’. 

\textbf{Imageability} is defined as the easiness or difficulty of constructing a mental image associated with that word, with 1 representing when the word is very difficult to imagine and 7 when it is very easy. It is easier to imagine something with words like ‘flag’ than with others like ‘charity’. 

\textbf{Context availability} is defined as the easiness or difficulty in associating that word with a context in which it could appear, with 1 representing when the word is very difficult to associate with a context and 7 when it is very easy. It is easier to construct sentences or search for usage examples for words like ‘table’ than for others like ‘citizenship’.

\textbf{Familiarity} is defined as the degree of familiarity, with 1 representing when the word is not very familiar and 7 otherwise. A word like ‘fish’ is more familiar than others like ‘quark’.  

In \parencite{stadthagen2017norms} the following variables are collected for 14031 words\footnote{All the variables have ranges between 1 (minimum) and 9 (maximum).}: 
\begin{itemize}
    \item \textit{Word}: word considered.  
    \item \textit{ValenceMean}: mean value of the valence for that word from the individual ratings given by the subjects.
    \item \textit{ArousalMean}: mean value of the arousal for that word from the individual ratings given by the subjects. 
    \item \textit{ValenceSD}: standard deviation of the valence values for that word from the individual ratings given by the subjects
    \item \textit{ArousalSD}: standard deviation of the arousal values for that word from the individual ratings given by the subjects
    \item \textit{\% ValenceRaters}: percentage of total subjects that have given a value to the valence. 
    \item \textit{\% ArousalRaters}: percentage of the total of subjects that has given a value to the arousal.
\end{itemize}

This last work is complemented with \parencite{stadthagen2018norms}, where the authors provide a larger lexicon (10491 words) that includes, together with valence and arousal variables (mean and deviation), the mean and standard deviation values for the following affective states: happiness, disgust, anger, fear, sadness. The authors also include a column called "Few\_Raters" that indicates whether the number of subjects used for that word is small or not, together with the dominant POS associated to that word.

Similarly, \parencite{hinojosa2016affective} also includes affective values for valence, arousal, happiness, disgust, anger, fear, sadness but for 875 words. This last lexicon also includes the value for concreteness.

In, the authors \parencite{alonso2015subjective} describe for 7040 words other characteristics such as the average age at which a word is usually learned (averageAoA, Age of Acquisition), the minimum (Min) and maximum (Max) age and the deviation in these age data (SD), as well as the literary frequency with which it is usually found. 

Finally, in \parencite{perez2021emopro} the authors provide the affective values for 1286 words. 
It includes some of the variables already seen, like the affectives states in terms of valence, arousal, happiness, disgust, anger, fear, sadness. Some of these words already appeared in other previous lexicons, like \parencite{stadthagen2018norms}, and the authors reuse the affective value from those sources. When that happens, they indicate it with a specific column (e.g. "Fear Source"). Some of the words are new and did not appear before. Along with these variables, they include new ones, like the AoA, and the dominant emotion associated to that word (e.g. "amar" (love) to "happiness"). They also include a new field called "emotionality" that indicates how much that word indicates an emotion. 

This are, to the best of our knowledge, the main lexicons for affective values of Spanish words that will serve as an equivalent to BAWL. We also consider these lexicons since the affective values associated to the words were obtained considering a general public from different ages, as opposed to more recent lexicons like \parencite{sabater2020spanish}, where the people involved were children and adolescents.

\subsection{Poetry and Psychology}
As we mentioned before, poetry contains an affective dimension that may evoke different sentiments, which can be quantified by inferring its GAM. But the affective dimension is not the only one present in a poem. Poems are also a way to express the psychological state of the author, as indicated in \parencite{czernianin2016poetry}. Here, the article shows how poetry is used as a way to discharge the mood of the authors. In fact, they analyse several poems to see how some of its content reflect psychological states such as suffering, happiness or hedonism. 
Following this, the psychological state of the author is reflected in the poem, and that also evokes a particular psychological state in the reader, as mentioned in \parencite{kao2012computational}. Here, the authors mention both how poetry is used as a way to explore and express emotions, as well as how it causes in the readers psychological states such as catharsis. In fact, \parencite{parastoo2016effect} conduct a study in which they analyse how reading poetry can be used as a therapy to treat psychotic patients. Thus, poetry can influence the reader's state to a point that it can even be used as a therapy to change or mitigate a particularly pernicious psychological state.
Complementing this, \parencite{shapiro2003can} show how including poetry within a medical student program enhances dimensions such as empathy, altruism, compassion, and caring toward patients.
The connection between psychology and emotions also appears in \parencite{jacobs2016elementary}, where the authors indicate how "psychological pleasure" is connected to the beauty of the text, which is often expressed through several emotions that are provoked in the reader. Indeed, in this work the authors conduct a research on Elementary Affective Decisions, and analyse how the decision process is influenced by basics emotions (e.g. happiness or disgust) within the context of Neurocognitive Poetics. 
Following this, \parencite{jacobs2019sentiment} shows that the connection between psychological and affective states can also appear within the characters of a literary work. Here, even though it is studied for prose texts, the authors model the affective state of characters (valence, arousal...), as well as the personality profile through a model inspired in the Big-5 (friendly, affectionate, hostile...).

Thus, it is interesting to know not only what affective states and sentiments are evoked by a poem (captured in the GAM), but also what psychological state the poem evokes, in order to contribute to its usage within all those contexts aforementioned. However, to the best of our knowledge there are no corpora that identify different groups of poems according to the psychological states that they reflect. Due to this, we find a research need in providing an annotated corpus of poems that identifies different subsets according to some psychological states, identified by tags.

Also, since poetry both evoke affective and psychological states intertwined, it is important to quantify how GAM changes according to the psychological state represented in its content.

\section{Methodology}
The methodology proposed consists of inferring the GAM of a sonnet based on the individual contribution of its words, and then validating that using a gold standard labelled corpus. Thus, we define an unsupervised approach to build the GAM and then we use domain knowledge to check it.

This Section first introduces the corpus included in this paper, Diachronic Spanish Sonnet Corpus with Psychological and Affective Labels, DISCO PAL. We begin by presenting the participants who annotated the corpus, and after that we will describe the corpus itself. We conclude introducing the methodology used, which includes the input data sources and the features built from them.

\subsection{Participants}
As mentioned before, the features were annotated by three experts in digital humanities, literature and linguistics, belonging to POSTDATA project.

The experts have annotated the sonnets independently (without knowing the annotations from the other experts) and following the same sonnet order. They did not know the author or the time period of the different sonnets; they only had access to the text itself. This was done in order to mitigate bias in their judgement. They used a csv file with rows containing the sonnet texts and columns with the different variables. Each of them assigned a value within the available range in the corresponding column. Those experts have individually annotated the same 274 sonnets for all the features described below.

\subsection{Materials}
DISCO PAL is a subset of a larger corpus, DISCO \parencite{ruiz2018disco} DISCO that consists of 4085 sonnets in Spanish language from the 15th to 19th century. From that corpus, in order to create DISCO PAL, the experts of POSTDATA have annotated a subset of 274 sonnets, with 184 belonging to the 19th century, 9 belonging to the 18th century and the other 81 belonging to the interval of 15th to 17th century. This is a relevant fact to consider because some sonnets are written in old Spanish, something that can significantly affect all the text mining analysis applied to the poems. Also, the number of authors used is 52, and from them only 3 are women (covering 12\% of the total sonnets).  With that, the corpus provided is rich, with many different authors belonging to different centuries, in line with the proposals of the scientific literature \parencite{eastman2015making}.

\begin{figure}[h!]
\centering
  \begin{tabular}{c@{\qquad}c@{\qquad}c}
  \includegraphics[width=0.99\columnwidth]{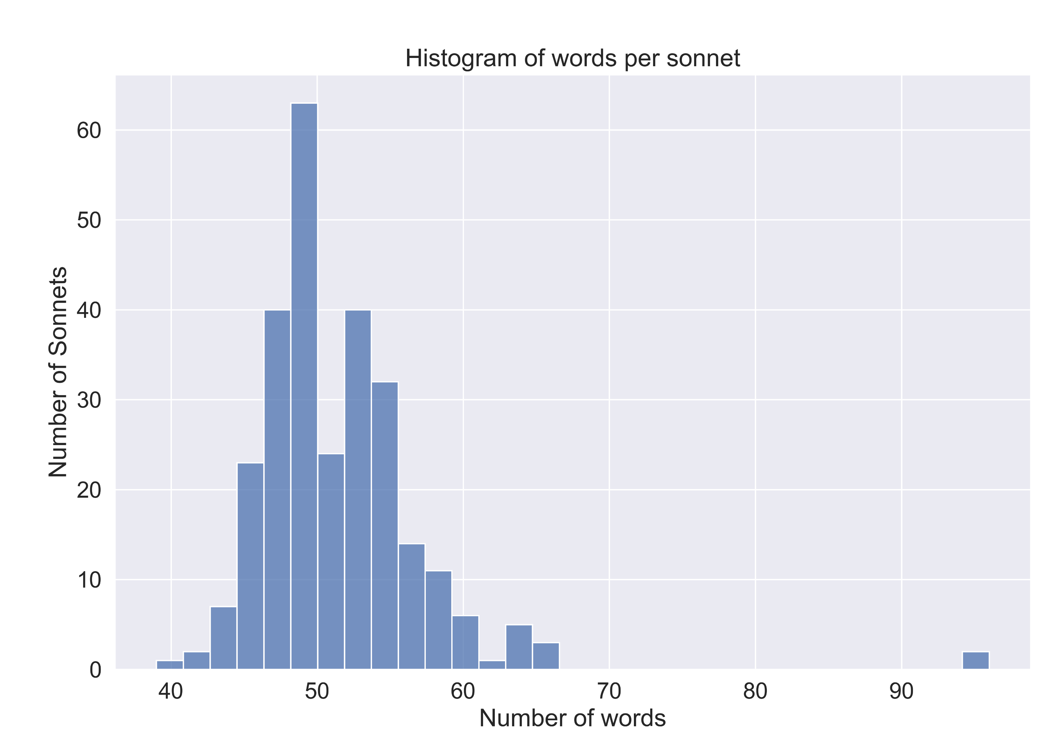}
  \end{tabular} 
  \caption{Histogram for the number of words per sonnet}
  \label{fig:histogram-words}
\end{figure}

The mean number of words per sonnet 51.6, the standard deviation is 5.9, and its associated histogram can be seen in Figure \ref{fig:histogram-words}.

There are three types of annotated features: affective, lexico-semantic and psychological. Affective features are detailed in Table \ref{table:affective-list} and have a range of 1 to 4, with 1 being the minimum value (the sonnet does not inspire very much that state) and 4 the maximum (the sonnet inspires it very much). The scale only uses integer values. The scale is the same for Lexico-sematic features, which are described in Table \ref{table:lexico-semantic-list}. Psychological features are binary and indicate whether the sonnet is related to that concept (1) or not (0). These features are described in Table \ref{table:psycho-list}.

\begin{table}[h!]
\centering
\begin{tabular}{@{}lll@{}} 
\toprule
& Affective features & \\ 
\midrule
 Valence & & Arousal \\
 Happiness & & Disgust\\ 
 Anger & & Sadness \\
 Fear & & \\ [1ex] 
\bottomrule
\end{tabular}
\caption{Affective features used with a scale of 1 to 4, using integer values.}
\label{table:affective-list}
\end{table}

\begin{table}[h!]
\centering
\begin{tabular}{@{}lll@{}} 
\toprule
& Lexico-semantic features & \\ 
\midrule
 Concreteness\footnote{Same as ‘Concreteness’ defined in the SOTA} & & Imageability\footnote{Same as ‘Imageability’ defined in the SOTA}\\
 Context availability\footnote{Same as ‘Context availability’ defined in the SOTA}  & & \\ [1ex] 
\bottomrule
\end{tabular}
\caption{Lexico-semantic features used with a scale of 1 to 4, using integer values.}
\label{table:lexico-semantic-list}
\end{table}

\begin{table}[h!]
\centering
\begin{tabular}{@{}lll@{}} 
\toprule
& Psychological features & \\ 
\midrule
Solitude (Soledad) & & Anxiety (Ansiedad)\\ 
Illusion (Ilusión) & & Anger/Wrath (Ira)\\
Daydream (Ensoñación) & & Instability (Inestabilidad)\\
Grandeur (Grandiosidad) & & Idealization (Idealización)\\
Pride (Orgullo) & &  Depression (Depresión)\\ 
Irritability (Irritabilidad) & & Disappointment (Desilusión)\\
Dramatisation (Dramatización) & & Prejudice (Prejuicio)\\
Aversion/Loathing (Aversión) & & Insecurity (Inseguridad)\\
Helplessness (Impotencia) & & Vulnerability (Vulnerabilidad)\\ 
Fear (Temor) & & Obsession (Obsesión)\\
Compulsion (Compulsión) & & \\[1ex] 
\bottomrule
\end{tabular}
\caption{Psychological features used with a scale of 0 to 1 (binary values).}
\label{table:psycho-list}
\end{table}

Regarding the psychological features, they were chosen considering their relevance in the literature \parencite{francotmf2016owl}. All these annotations can be used for calculating different metrics in the recovery of poems (such as precision, for example). 

We show below a sonnet example along with its English translation and the median annotations by the experts:

\resizebox{160pt}{!}{%
\begin{tabular}{@{}lll@{}}
\noindent
\\
\\
Raro Fénix de Amor, que en vivas llamas,\\
esplendor inmortal tienes logrado,\\
leños de aroma son, los que has juntado\\
en olor de virtudes que derramas.\\
\\
Alta Hoguera te eriges, que así amas\\
afectos recogiendo enamorado,\\
que el Pecho, en sacro amor, todo abrasado,\\
hoguera es elevada, en que te inflamas.\\
\\
A rayos del Sol Cristo, Ave lucida,\\
del corazón las alas, velozmente\\
bates, por verte en fuego renacida.\\
\\
Fénix te considero, en Pira ardiente,\\
que él en su muerte nace a nueva vida,\\
y es tu Ocaso en la Tierra, al Cielo, Oriente.\\
\\
\end{tabular}%
}
\quad
\resizebox{170pt}{!}{%
\begin{tabular}{@{}lll@{}}
\\
\\
Rare Phoenix of Love, that in living flames,\\
immortal splendor you have achieved,\\
aroma logs are the ones you have gathered\\
in the smell of virtues that you pour out.\\
\\
High Bonfire you erect, that is how you love\\
collecting affections in love,\\
that the chest, in sacred love, all burned,\\
bonfire is high, in which you get inflamed.\\
\\
In the rays from Sun Christ, magnificent bird,\\
from the heart the wings, swiftly\\
you bat, to see you in fire reborn.\\
\\
Phoenix I consider you, in Burning Pyre,\\
that in his death he is born to new life,\\
and it's your Sunset on Earth, to Heaven, East.\\
\\
\end{tabular}%
}

This sonnet (sonnet no. 18 in the corpus) belongs to Juan de Aguilar (S. XIX). The median annotations between the 3 annotators appear in Table \ref{table:example-annotations}.

\begin{table}[h!]
\centering
\begin{tabular}{@{}lllll@{}}
\toprule
\textbf{Feature} & \textbf{Median value} & \textbf{} & \textbf{Feature} & \textbf{Median value} \\ \midrule
happiness      & 4 &  & Helplessness          & 0 \\
sadness        & 1 &  & Vulnerability        & 0 \\
Solitude       & 0 &  & Fear (binary)        & 0 \\
Illusion       & 0 &  & Obsession            & 0 \\
Daydream       & 0 &  & Compulsion           & 0 \\
Grandeur       & 1 &  & Prejudice            & 0 \\
Pride          & 0 &  & Dramatisation        & 1 \\
Irritability   & 0 &  & valence              & 3 \\
Anxiety        & 0 &  & arousal              & 3 \\
Anger          & 0 &  & anger                & 1 \\
Instability    & 0 &  & Fear (ordinal)       & 1 \\
Idealization   & 1 &  & disgust              & 1 \\
Depression     & 0 &  & concreteness         & 3 \\
Disappointment & 0 &  & imageability         & 3 \\
Aversion       & 0 &  & context availability & 3 \\
Insecurity     & 0 &  &                      &   \\ \bottomrule
\end{tabular}
\caption{Example of annotations for a sonnet.}
\label{table:example-annotations}
\end{table}

\subsection{Procedure}
The methodology is divided into two parts. First, we build the GAM values from the individual words within the sonnets. This is done by using several external lexicons to assign an affective or lexico-semantic values to the individual Spanish. These lexicons include some of the ones already introduced in the Related Work Section. We use: \parencite{hinojosa2016affective}, \parencite{ferre2017moved}, \parencite{guasch2016spanish}, \parencite{stadthagen2018norms} and \parencite{perez2021emopro}. These lexicons will be combined into one lexicon. When there are several possible values for the same word, we will use the median value between them. We use those affective or lexico-semantic values at a word level and aggregate them to model the GAM for the whole sonnet. This is accomplished by generating different features, as indicated below:
\begin{itemize}
    \item \textit{valence\_mean}: mean of valence mean values for the individual words.
    \item \textit{valence\_sd}: mean of valence standard deviation values for the individual words.
    \item \textit{arousal\_mean}: mean of arousal mean values for the individual words.
    \item \textit{arousal\_sd}: mean of arousal standard deviation values for the individual words.
    \item \textit{happiness\_mean}: mean of happiness mean values for the individual words.
    \item \textit{happiness\_sd}: mean of happiness standard deviation values for the individual words.
    \item \textit{anger\_mean}: mean of anger mean values for the individual words.
    \item \textit{anger\_sd}: mean of anger standard deviation values for the individual words.
    \item \textit{sadness\_mean}: mean of sadness mean values for the individual words.
    \item \textit{sadness\_sd}: mean of sadness standard deviation values for the individual words.
    \item \textit{fear\_mean}: mean of fear mean values for the individual words.
    \item \textit{fear\_sd}: mean of fear standard deviation values for the individual words.
    \item \textit{disgust\_mean}: mean of disgust mean values for the individual words.
    \item \textit{disgust\_sd}: mean of disgust standard deviation values for the individual words.
    \item \textit{concreteness\_mean}: mean of concreteness mean values for the individual words.
    \item \textit{concreteness\_sd}: mean of concreteness standard deviation values for the individual words.
    \item \textit{imageability\_mean}: mean of imageability mean values for the individual words.
    \item \textit{imageability\_sd}: mean of imageability standard deviation values for the individual words.
    \item \textit{cont\_ava\_mean}: mean of context availability mean values for the individual words.
    \item \textit{cont\_ava\_sd}: mean of context availability standard deviation values for the individual words.
    \item \textit{max\_arousal}: maximum value of arousal mean values for the individual words.
    \item \textit{min\_arousal}: minimum value of arousal mean values for the individual words.
    \item \textit{max\_valence}: maximum value of valence mean values for the individual words.
    \item \textit{min\_valence}: maximum value of valence mean values for the individual words.
    \item \textit{arousal\_span}: $max\_arousal - min\_arousal$
    \item \textit{valence\_span}: $max\_valence - min\_valence$
    \item \textit{CorAro}: Spearman's correlation between the arousal mean value of the words and their position in the sonnet
    \item \textit{CorVal}: Spearman's correlation between the valence mean value of the words and their position in the sonnet
    \item \textit{AbsCorAro}: absolute value of CorAro
    \item \textit{AbsCorVal}: absolute value of CorVal
    \item \textit{sigma\_aro}: $\frac{arousal\_mean}{1/\sqrt{N}}$ with N the number of words in the sonnet.
    \item \textit{sigma\_val}: $\frac{valence\_mean}{1/\sqrt{N}}$ with N the number of words in the sonnet.
\end{itemize}

Then, we define the features to be annotated in the DISCO PAL sonnet corpus in order to analyse the quality of the inferred GAM values. These features were the ones described in the Materials Section. Thus, we will compare every feature associated to anger, sadness, disgust, arousal, valence, concreteness, imageability and context availability to the value annotated by the experts. We will also analyse these comparisons considering the psychological tags. For that, we will consider separately sonnets that belong to a particular psychological category and analyse the GAM against the annotated values of the affective or lexico-semantic features for that subset of sonnets only.

\section{Evaluation}
The evaluation steps are the following ones:
\begin{itemize}
    \item Study the reliability of the DISCO PAL corpus annotated by the POSTDATA experts. 
    \begin{itemize}
        \item We analyse the bivariate correlation between annotators for the affective features in order to see if they are logical. For instance, "valence" in one of the annotators should be positively correlated with the value of "valence" from another expert. Also, a variable like "valence" should also be positively correlated with another one that represents positive affective states, such as "happiness".
        \item Then, we check the level of agreement between the annotators in order to see if there are significant discrepancies between them. If the level of agreement is enough, we can proceed to the next point.
    \end{itemize}
    \item Analyse the relationship between the features annotated by the experts and the ones obtained through the GAM infer methodology shown before. This analysis is carried out at three levels.
    \begin{itemize}
        \item First, we analyse the bivariate correlation between the GAM inferred features and their annotated counterpart. We check if that correlation is over a minimum threshold. Literature \parencite{schober2018correlation} indicates a basic reference of [0.1-0.39] as a weak correlation, [0.4-0.69] as a moderate correlation, [0.7-0.89] strong correlation, and >0.9 very strong correlation.
        \item Then, we analyse the partial correlation between those same GAM inferred features and the annotated ones. This is done by building a regression model over the GAM features (independent variables) and each label at a time (dependent feature). We check the level of significance using the p-value for the inferred feature, the r-squared value, and the feature coefficient.
        \item The previous analysis is done considering all the annotated DISCO PAL corpus, as well as separating it by the annotated psychological categories. We want to see if the results differ significantly. 
        \item Finally, in order to analyse differences in the GAM depending on the psychological category, we perform a One-way ANOVA hypothesis contrast. We compare the mean values for a particular GAM inferred feature between the subset that belongs to a specific psychological category against the other subset of sonnets. There will be differences if the p-value is less than 0.05.
    \end{itemize}
\end{itemize}

\subsection{Supplementary material}
The materials included in this article are three csv files with the annotations made by the experts, as well as a csv file with metadata information about the annotated sonnets. This metadata csv is included in order to allow the reference between the DISCO PAL and the original source DISCO. The fields included in the metadata csv are:
\begin{itemize}
    \item author: author of the sonnet.
    \item year: year or century of publication.
    \item title: title of the sonnet.
    \item \textit{id\textunderscore sonnet}: unique id used by DISCO for that sonnet.
    \item \textit{file\textunderscore path}: file name path to that sonnet in the per-sonnet folder in DISCO.
\end{itemize}

We also include a csv with the aggregated annotations by the three experts.
All data provided is located at \parencite{barbado2019pal}.

\subsection{Reliability and validity of DISCO PAL corpus}
\begin{figure}[h]
\centering
  \begin{tabular}{c@{\qquad}c@{\qquad}c}
  \includegraphics[width=1\columnwidth]{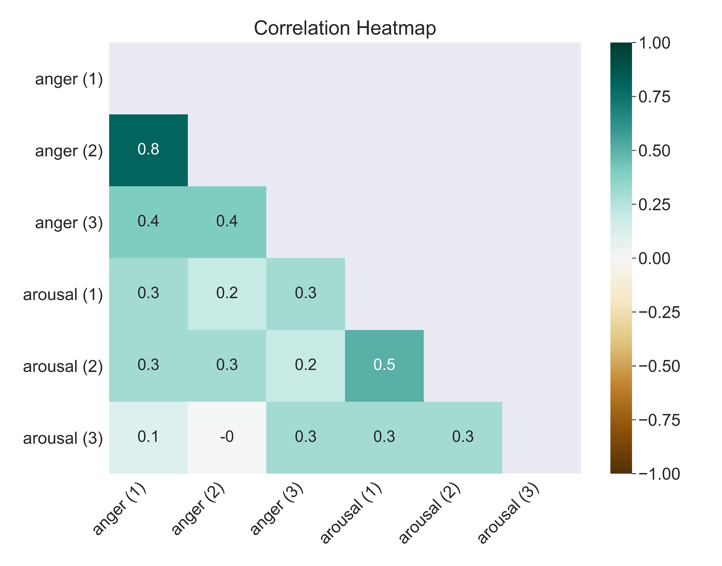}
  \end{tabular} 
  \caption{Bivariate correlations between the annotators for arousal and related affective states (anger)}
  \label{fig:corr_ann_arousal}
\end{figure}

\begin{figure}[h]
\centering
  \begin{tabular}{c@{\qquad}c@{\qquad}c}
  \includegraphics[width=1\columnwidth]{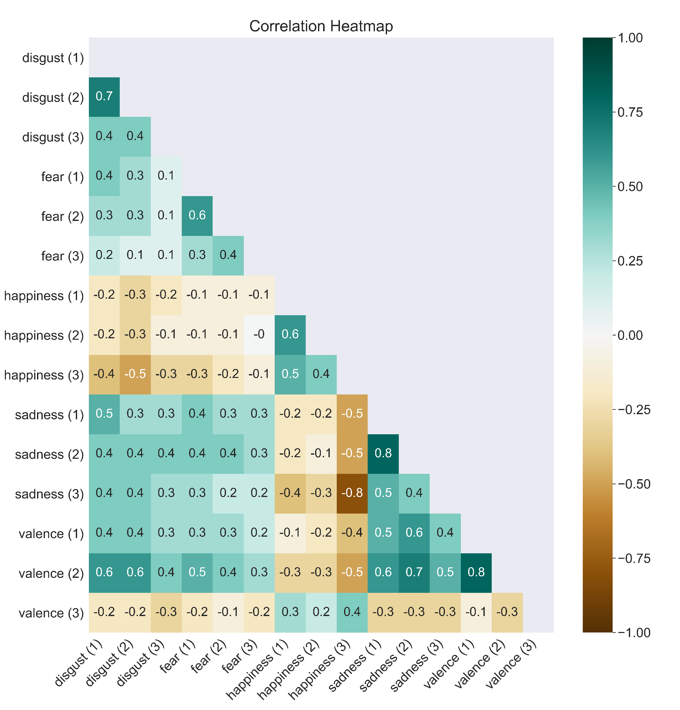}
  \end{tabular} 
  \caption{Bivariate correlations between the annotators for arousal and related affective states (happiness, sadness, fear, disgust)}
  \label{fig:corr_ann_valence}
\end{figure}

The first approach to study the reliability and validity of DISCO PAL is to see that the correlations between annotators and between specific features are logical. In Figure \ref{fig:corr_ann_arousal} we see the correlations for two features, "arousal" and "anger", considering the three annotators. We see how these two features are positively correlated. Both "arousal" and "anger" have positive correlations between the annotators. They also have positive correlations when the values are compared for the same annotator.
In Figure \ref{fig:corr_ann_valence} we see the correlations for "valence", "disgust", "fear", "happiness" and "sadness". For annotators 1 and 2, we see how "valence" is negatively correlated with that same feature in annotator 3. We also see how it is negatively correlated with "happiness", while having a positive correlation with the remaining features. This indicates that annotators 1 and 2 have used a reversed scale in this feature. Because of that, for further analyses, we will reverse their results for the "valence" feature. For annotator 3 we see that the correlations are correct. The remaining features also seem to have logical correlations (e.g. positive for "disgust" and "fear", negative for "happiness" and "fear").  

The second step is analyzing the agreement between the three annotators. This is accomplished by obtaining the Krippenndorff Alpha \parencite{krippendorff2011computing}, or k-alpha, between the annotations made by the 3 experts for each of the features. 

K-alpha is a metric that generalizes other metrics that are responsible for quantifying the reliability between annotators (inter-rater reliability). It can be used for both ordinal and nominal annotations, as well as with any number of annotators. K-alpha yields a value between 0 and 1, where 1 represents full agreement. However, there are different criteria regarding when to consider that there is enough agreement between annotators. If the acceptance criteria is strict, only expert annotations are accepted as truly valid if there is a k-alpha of at least 0.8 \parencite{carletta1996assessing}. Other laxer criteria set the minimum at 0.21, defining the following thresholds \parencite{landis1977measurement}:

\begin{itemize}
    \item $K < 0$: Very low
    \item $0 < K < 2$: Light
    \item $0.21 < K < 0.4$: Acceptable
    \item $0.41 < K < 0.60$: Moderate
    \item $0.61 < K < 0.80$: Substantial
    \item $0.81 < K < 1$: Perfect
\end{itemize}

The k-alpha results considering the three annotators together are shown in Table \ref{table:k-alpha}. "k 12" represents the agreement between annotators 1 and 2 (analogous to "k 13" and "k 23"). In bold we see the k-alpha values that are below the "acceptable" threshold. Between most of the features and annotators, the level of agreement is above $K >= 0.21$, with some cases that reach the "substantial" level ($K >= 0.61$). However, there are combinations that yield levels below that 0.21 threshold, particularly for some features from annotator 3 when compared to either annotator 2 or annotator 1. 
For annotators 1 and 2 all the features are validated. For annotators 1 and 3, and for annotators 2 and 3, the same 7 features have a k-alpha below 0.21 (with an additional feature for the case of 1 and 3). 
Considering together all the annotators, the k-alpha results are all above 0.21, with the exception of the feature "happiness" (as seen in "k all" column). This means that 97\% of the features have a moderate level of agreement (or better).

In order to conduct further analyses, those three annotation sets should be combined into only one vector. A proposal for doing it is by using the median value between the values of the three experts. In that way, if there is a discrepancy between two annotators and a third one, the final value used will be the one that agrees with most of them. This median value will act as a proxy "annotator" than agrees with the three experts. Indeed, as also shown in Table \ref{table:k-alpha}, the agreement versus each annotator is very high. It is still above 0.21 for annotators 1 and 2, while reducing the discrepancy against annotator 3 in 4 features out of 8. Some of the annotators left with nulls some of the psychological tags (annotator 1 left 38 sonnets with an average of 3.6 psychological tags without annotations, annotator 2 left 4, and annotator 3 left 1). For building this proxy annotator, we filled those missing categories with 0, assuming that if the concept was not explicit by an annotators, it does not appear. This only applies for cases where there is one sonnet-psychological tag missing for only one of the annotators, but not for the other two. The affective and lexico-semantic features are fully annotated in all the sonnets.

Using this proxy annotator, we get a number of sonnets per psychological category as shown in Table \ref{table:sonnets-per-psycho}. We see how the categories "Obsession" and "Prejudice" are the ones that appear in less sonnets.

\begin{table}[h!]
\centering
\begin{tabular}{@{}ll@{}}
\toprule
\textbf{Category} & \textbf{Number of sonnets} \\ \midrule
all            & 274 \\
Anxiety        & 76  \\
Aversion       & 99  \\
Depression     & 39  \\
Disappointment & 47  \\
Dramatisation  & 108 \\
Illusion       & 73  \\
Helplessness    & 62  \\
Instability    & 64  \\
Insecurity     & 44  \\
Anger          & 57  \\
Obsession      & 32  \\
Pride          & 72  \\
Prejudice      & 30  \\
Fear (binary)  & 94  \\
Vulnerability  & 129 \\
Compulsion     & 56  \\
Daydream       & 46  \\
Grandeur       & 105 \\
Idealization   & 107 \\
Irritability   & 36  \\
Solitude       & 63  \\ \bottomrule
\end{tabular}
\caption{Sonnets per psychological category.}
\label{table:sonnets-per-psycho}
\end{table}

In Table \ref{table:sonnets-per-psycho} we see that "Prejudice" and "Obsession" are the categories with fewer sonnets (30 and 32 respectively). Although this number is smaller when compared to other categories (e.g. "Dramatisation" has 108 sonnets), it is enough from a statistical point of view based on a power analysis with an alpha of 0.05, a Cohen’s d of 0.8 and the default statistical power of 0.8 (which sets the minimum in 26) \parencite{cohen1992power, sullivan2012using}. The number of sonnets per category is also higher when compared to other similar analyses, such as \parencite{ullrich2017relation, aryani2016measuring}, where the authors work with the categories “friendliness”, “sadness” and “spitefulness” and they are associated to 19, 21 and 17 poems respectively.

\subsection{Analysis of DISCO PAL corpus for individual affective word modelling}
As previously mentioned, the original corpus consists of 4085 sonnets in Castilian language from 15th to 19th century, collected from the corpus DISCO from POSTDATA (UNED). From there, we have selected 274 sonnets, which have been annotated with specific affective features, inspired by the literature, in particular \parencite{ullrich2017relation}. 

That article indicates how they modeled the GAM for the poems by using the BAWL lexicon as input source. This lexicon contains 6000 words in German. In order to associate the value of features to individual words (for modeling the GAM later on), they use the different words available in the poems. To increase the number of words that match the entries in these tables, the words of the poems are lemmatized, and stopwords are removed. In this way, it is possible to find a match for 90\% of the words that appear in the poems with a word within the BAWL lexicon. The remaining 10\% of words that do not appear in these tables are usually proper names. 

With that, we consider in this paper how many words from DISCO PAL appear in input lexicons used for assigning the affective or lexico-semantic value to the individual words.

Table \ref{table:words-per-category} shows how many unique words are considering all the sonnets within DISCO PAL, as well as in each of the subsets associated to a psychological category. It also shows how many unique words are when they are lemmatized or stemmed (with SnowBall stemming algorithm \parencite{porter2001snowball}). As expected, both lemmatization and stemming techniques reduce the number of words (with stemming reducing them more than the lemmatization).

\begin{table}[h!]
\centering
\begin{tabular}{@{}llll@{}}
\toprule
\textbf{category} & \textbf{n words} & \textbf{n words stem} & \textbf{n words lem} \\ \midrule
all            & 5898    & 3651         & 4613        \\
Anxiety        & 2278    & 1690         & 1927        \\
Aversion       & 2846    & 2054         & 2356        \\
Depression     & 1352    & 1080         & 1198        \\
Disappointment & 1624    & 1284         & 1395        \\
Dramatisation  & 3055    & 2159         & 2509        \\
Illusion       & 2261    & 1682         & 1904        \\
Helplessness    & 1987    & 1484         & 1668        \\
Instability    & 2003    & 1505         & 1703        \\
Insecurity     & 1492    & 1184         & 1297        \\
Anger          & 1904    & 1497         & 1640        \\
Obsession      & 1170    & 955          & 1025        \\
Pride          & 2264    & 1725         & 1934        \\
Prejudice      & 1181    & 995          & 1067        \\
Fear (binary)  & 2756    & 1990         & 2291        \\
Vulnerability  & 3319    & 2286         & 2724        \\
Compulsion     & 1821    & 1412         & 1566        \\
Daydream       & 1636    & 1277         & 1400        \\
Grandeur       & 3076    & 2200         & 2550        \\
Idealization   & 3040    & 2166         & 2514        \\
Irritability   & 1308    & 1086         & 1164        \\
Solitude       & 1978    & 1518         & 1694        \\
\bottomrule
\end{tabular}
\caption{Number of words (original, after lemmatization or stemming) in the original DISCO PAL corpus (total and per psychological category)}
\label{table:words-per-category}
\end{table}

Table \ref{table:per-words-per-corpora} shows the words of the DISCO PAL corpus that match the ones in the different source lexicons, and Table \ref{table:fraction-words} shows the same but when the words from DISCO PAL, as well as the words from the source lexicons, are either stemmed or lemmatized. It can be seen that using lemmatization or stemming techniques improves, as expected, the number of words that match the input lexicons. Since stemming improves the matching, we will use this technique for the subsequent analyses. Lemmatization and stemming scenarios also include the elimination of stopwords. 

\begin{table}[h!]
\centering
\begin{tabular}{@{}lllllll@{}}
categories & \rot{all lexicons} & \rot{\cite{perez2021emopro}} & \rot{\cite{stadthagen2018norms}} & \rot{\cite{hinojosa2016affective}} & \rot{\cite{guasch2016spanish}} & \rot{\cite{ferre2017moved}} \\ \midrule
all            & 0.37 & 0.05 & 0.22 & 0.03 & 0.06 & 0.11 \\
Anxiety        & 0.46 & 0.07 & 0.26 & 0.04 & 0.09 & 0.16 \\
Aversion       & 0.44 & 0.06 & 0.25 & 0.04 & 0.08 & 0.14 \\
Depression     & 0.48 & 0.08 & 0.26 & 0.05 & 0.09 & 0.17 \\
Disappointment & 0.48 & 0.07 & 0.26 & 0.05 & 0.1  & 0.17 \\
Dramatisation  & 0.42 & 0.07 & 0.23 & 0.04 & 0.08 & 0.13 \\
Illusion       & 0.44 & 0.06 & 0.25 & 0.04 & 0.08 & 0.14 \\
Helplessness    & 0.45 & 0.07 & 0.24 & 0.04 & 0.09 & 0.16 \\
Instability    & 0.46 & 0.07 & 0.25 & 0.04 & 0.09 & 0.16 \\
Insecurity     & 0.5  & 0.09 & 0.27 & 0.05 & 0.1  & 0.17 \\
Anger          & 0.47 & 0.07 & 0.26 & 0.04 & 0.1  & 0.17 \\
Obsession      & 0.47 & 0.08 & 0.26 & 0.05 & 0.09 & 0.16 \\
Pride          & 0.46 & 0.06 & 0.26 & 0.05 & 0.09 & 0.14 \\
Prejudice      & 0.48 & 0.06 & 0.25 & 0.05 & 0.09 & 0.17 \\
Fear (binary)  & 0.44 & 0.07 & 0.25 & 0.04 & 0.08 & 0.14 \\
Vulnerability  & 0.42 & 0.06 & 0.23 & 0.04 & 0.08 & 0.13 \\
Compulsion     & 0.47 & 0.07 & 0.26 & 0.05 & 0.09 & 0.15 \\
Daydream       & 0.45 & 0.06 & 0.26 & 0.05 & 0.09 & 0.14 \\
Grandeur       & 0.42 & 0.06 & 0.24 & 0.04 & 0.08 & 0.13 \\
Idealization   & 0.42 & 0.06 & 0.24 & 0.04 & 0.08 & 0.14 \\
Irritability   & 0.5  & 0.08 & 0.27 & 0.05 & 0.1  & 0.18 \\
Solitude       & 0.46 & 0.07 & 0.26 & 0.04 & 0.09 & 0.15 \\ \bottomrule
\end{tabular}
\caption{Fraction of words (original) from the original DISCO PAL corpus (total and per psychological category) in the different source lexicons}
\label{table:per-words-per-corpora}
\end{table}

Tables \ref{table:per-words-per-corpora} and \ref{table:fraction-words} show that when all the lexicons are combined together, the percentage of matching words increases, and, for lemmatization and stemming, is above 50\% for both all DISCO PAL, as well as for the individual psychological categories.

Considering lemmatization, the percentage of matching words is 56\% (when all the source lexicons are combined) versus the 90\% from \parencite{ullrich2017relation}. Table \ref{table:top-missing-words} shows the top 14 most common missing words from the DISCO PAL corpus in all of the source lexicons.

\begin{table}[h!]
\centering
\begin{tabular}{@{}ll@{}}
\toprule
\textbf{Word} & \textbf{Number of occurrences} \\ \midrule
viva        & 12    \\
soberano    & 12    \\
impía       & 11    \\
porfío      & 11    \\
españa      & 10    \\
contigo     & 10    \\
ruego       & 10    \\
verte       & 9     \\
apolo       & 9     \\
planta      & 8     \\
beldad      & 7     \\
ceniza      & 7     \\
excelso     & 7     \\
eternamente & 7    \\ \bottomrule
\end{tabular}
\caption{Most common words in DISCO PAL corpus missing in the source lexicons (excluding stopwords). It shows how many times that word appears.}
\label{table:top-missing-words}
\end{table}

As shown in Table \ref{table:top-missing-words}, most of the common missing words represent archaic terms that are not not frequently used now (e.g. 'impía', 'porfío', 'beldad'...). It also includes some proper nouns (p.e. 'apolo' or 'españa').

This scenario will probably hinder the results from the GAM in comparison to \parencite{ullrich2017relation} since there are more absent words in the source lexicons, even after removing stopwords and performing stemming (where the matching percentage is 68\%, still below 90\%).

\subsection{GAM analysis}
Following a similar approach to \parencite{ullrich2017relation}, the source lexicons mentioned are going to be used as an input source in order to infer the GAM value of the sonnets. The results are going to be validated against the annotated values by the POSTDATA experts. 

As mentioned previously, the evaluation is going to be assessed against the median value derived from the three annotators. The words from the source lexicons, as well as the words from DISCO PAL, are stemmed. Using the source lexicons, we build the features described in the Methodology Section for each of the sonnets. Since a stemmed word can appear multiple times in the source lexicons (p.e. "bees" and "bee" will be the same word after stemming), the final value assigned to that word is the average between all the words with the same stem. These features represent the inferred GAM for that sonnet. 

\begin{figure}[h]
\centering
  \begin{tabular}{c@{\qquad}c@{\qquad}c}
  \includegraphics[width=1\columnwidth]{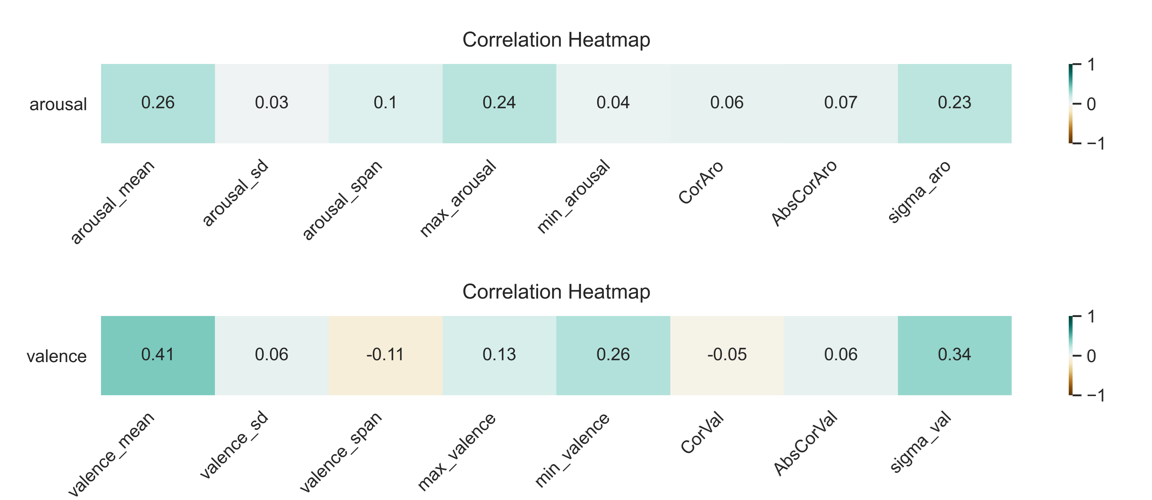}
  \end{tabular} 
  \caption{Bivariate correlations between the GAM features (arousal and valence) from source lexicons and the median annotator labels. Y axis annotated feature, X axis inferred ones. \label{fig:corr_aro_val}}
\end{figure}

Considering that, Figure \ref{fig:corr_aro_val} shows the Spearman's bivariate correlation between the annotated feature and their inferred GAM counterparts. We see that "arousal\_mean" has a significant correlation (albeit a weak one). "valence\_mean" has a moderate correlation. There are other features, such as "max\_arousal", "sigma\_aro", "min\_valence", and "sigma\_val" that also have weak correlations.

\begin{figure}[h]
\centering
  \begin{tabular}{c@{\qquad}c@{\qquad}c}
  \includegraphics[width=1\columnwidth]{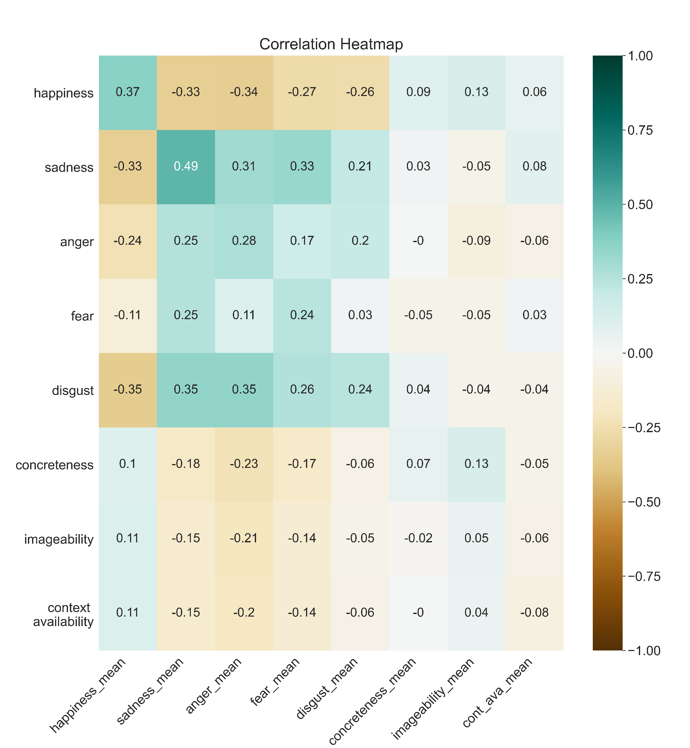}
  \end{tabular} 
  \caption{Bivariate correlations between the GAM features (affective) from source lexicons and the median annotator labels. Y axis annotated feature, X axis inferred ones.\label{fig:corr_affective}}
\end{figure}

Figure \ref{fig:corr_affective} shows the bivariate correlations for the remaining features. 5 out of the 8 features inferred have significant correlations with respect to their annotated counterparts. For "sadness", the correlation is moderate, and for "happiness", "anger", "fear" and "disgust", the correlation is weak (though close to moderate). It is also interesting to see that some inferred GAM features have significant correlations to other annotated features that could be expected. This is the case of "sadness\_mean", "anger\_mean", "fear\_mean" and "disgust\_mean". They all have significant and positive correlations when compared to features annotated like "fear", "anger", "sadness" or "disgust". They also have significant and negative correlations when compared to "happiness". 
The remaining features ("concreteness", "imageability" and "context availability") do not yield significant bivariate correlations (they are all below 0.1). Thus, the lexico-semantic features are the ones that have a lower correlation.

Following this analysis, Tables \ref{table:partial-analysis-all} and \ref{table:partial-analysis-per-psycho-tag} show the partial dependence between each GAM feature and their counterpart annotated by the experts. As mentioned before, a linear regression model is trained over all sonnets, using all GAM features as independent variables, and using one of the annotated features as dependent variable. Then, we get the p-value of the corresponding GAM feature, and see if that value is relevant, using a threshold of 0.05. ($p<0.05$ meaning it is significant). When the p-value if below 0.05, we indicate if in the "sign" column with a "yes".
We also check the coefficient of that feature to see that it is $>0$ (a negative coefficient would mean that even if the model is fitted properly, the relationship between both features is not coherent). We also include the adjusted r-squared value in order to see if the model is well-fitted.
In Table \ref{table:partial-analysis-all} we see the results for the whole DISCO PAL corpus. All the features have a p-value below the threshold, a positive coefficient while also having a high adjusted r-squared. In Table \ref{table:partial-analysis-per-psycho-tag}, we also see that in all the subset of sonnets belonging to each of the psychological categories, the features' coefficient is positive and the features are significant for predicting the corresponding annotated value. This indicates that GAM could be inferred whether we use the whole corpus or only a subset for a specific psychological category. 

\begin{table}[h!]
\centering
\resizebox{\textwidth}{!}{%
\begin{tabular}{@{}lllllll@{}}
\toprule
\textbf{Category} & \textbf{Feature (annotated)} & \textbf{Feature (GAM)} & \textbf{r2} & \textbf{coeff} & \textbf{p-value} & \textbf{sign} \\ \midrule
all & valence              & valence\_mean      & 0.92 &  1.71 & $<$ 0.001 & yes \\
all & arousal              & arousal\_mean      & 0.9  &  2.1  & $<$ 0.001 & yes \\
all & happiness            & happiness\_mean    & 0.8  &  1.28 & $<$ 0.001 & yes \\
all & anger                & anger\_mean        & 0.79 &  0.89 & $<$ 0.001 & yes \\
all & sadness              & sadness\_mean      & 0.85 &  0.68 & $<$ 0.001 & yes \\
all & fear                 & fear\_mean         & 0.84 &  1.08 & $<$ 0.001 & yes \\
all & disgust              & disgust\_mean      & 0.82 &  0.7  & $<$ 0.001 & yes \\
all & concreteness         & concreteness\_mean & 0.8  &  1.69 & $<$ 0.001 & yes \\
all & imageability         & imageability\_mean & 0.78 &  1.84 & $<$ 0.001 & yes \\
all & context availability & cont\_ava\_mean    & 0.78 &  2.03 & $<$ 0.001 & yes \\ \bottomrule
\end{tabular}%
}
\caption{Partial dependence analysis between annotated features and their inferred GAM counterpart (for all sonnets)}
\label{table:partial-analysis-all}
\end{table}

If we compare the results of our GAM extraction process against the ones on \parencite{ullrich2017relation}, we need to focus on the subset of features that appear in both of the papers. Those features are Valence and Arousal. Thus, we can compare the annotated GAM value for those features against their inferred counterparts, as well as to other features related to them, like CorAro or ValenceSpan. For Valence, the bivariate correlation of the inferred Valence value in \parencite{ullrich2017relation} is 0.65. For Arousal is 0.54. In both cases the partial correlation analysis shows statistical significance while using those inferred features as predictors for the annotated one. In our case, the bivariate correlation values are smaller, but they are still significant. This is also true for some other related features, as previously mentioned. Regarding the partial dependence analysis, we have validated those features.

Finally, we analyse if there are significant differences in the GAM (using the annotated value) between subsets depending on whether they refer to a specific psychological label or not. We perform a One-way ANOVA hypothesis contrast for each combination between a GAM feature and psychological label. The results for those combination that had p-values less than 0.05 are included in Table \ref{table:oneway-anova}. That table also includes the mean value for the GAM considering the sonnets annotated with that psychological tag, M (=1), and the other ones, M (=0). As we can see, from among the 210 possible combinations, 127 of them yielded significant differences in the GAM depending on the subset considered. 

\subsection{Limitations of our Approach}
There are several limitations within our approach. The first one is that the analysis is limited to the size of 274 sonnets from DISCO PAL. It would be interesting to perform it over a bigger corpus of annotated sonnets. However, as we mentioned earlier, we think that the corpus size is big enough for carrying out these analyses and obtain statistically meaningful results, since our corpus is has more poems per independent category than other corpora from the literature (\cite{ullrich2017relation}, \cite{aryani2016measuring}, \cite{jacobs2017s}, \cite{obermeier2013aesthetic}, \cite{haider2020po}). Also, the threshold found using the statistical power serves as another reason to support the statistical analyses carried out.

Another limitation is that the analysis is applied only over a group of sonnets in Castilian from the 15th to 19th century. Those sonnets contain many archaic words, and that reduces the matching between them against the source lexicons used to assign the affective or lexico-semantic value for individual words. In fact, the ratio of words in those lexicons, as already mentioned, is lower than other analyses within the literature, influenced in part by this aspect.

Also, though there is an acceptable agreement between the annotators for most of the features, the agreement is not perfect. This is something that also influences the results obtained.

Finally, there could be a possible bias due to the fact that the expert annotators have a profile specialized in digital humanities. If the annotators were experts in psychology, for instance, the results may differ.

\section{Conclusion and Future Work}
This Section concludes with a final reflection based on the results of the analyses carried out, as well as indicating possible lines of research that can be pursued.

\subsection{Conclusions}
This article presents a methodology to infer GAM feature values for Spanish poetry, using available lexicons that contains affective or lexico-semantic feature values for individual words. This GAM methodology is unsupervised, needing no prior information about the sonnets themselves.

The proposal is evaluated using a subset of sonnets annotated by domain experts. This article includes a corpus of 274 sonnets with features annotated. The sonnets are from Spanish authors from different time periods (from 15th to 19th century).
These sonnets are annotated using both affective or lexico-semantic features that indicate the intensity level of that feature within the sonnet, and concepts that belong to the psychological domain, indicating whether a sonnet content is related to that concept or not. They were annotated by three domain experts who belong to POSTDATA project (UNED). This corpus is shared as part of this article.

Then, we conduct an analysis on the level of agreement of the features annotated by the three experts. The result is that at least 97\% of the features have an adequate level of agreement. The results improve when we use the median value for the three annotators.

Using the median vector, we validate that it is feasible to model the GAM of a sonnet through several affective or lexico-semantic features built from their individual words. This is checked by analysing the bivariate correlation, as well as the partial dependence, between those features and their annotated counterparts. The results are particularly good for valence, arousal, happiness, sadness, fear, anger and disgust.

Finally, after considering results for all the sonnets together, we analyse if the GAM modelled for each of the subset of sonnets that belong to the different psychological categories differ significantly. We saw significant differences for some features and some psychological categories between the GAM of the sonnets that belong to it and the remaining ones.

\subsection{Future Work}
This subsection details the possible lines of research that can be pursued following the results presented in this article. There are two main group of research lines that are considered at this point. One is related to the improvement of the data quality involved in the GAM methodology, and the other is related to the applications of the DISCO PAL corpus.

Related to the data quality research areas, there are two fields of improvement. First, all the source lexicons used for the feature values of the individual words lack many of archaic words that are present in the sonnets. It would be useful to enrich those lexicons with these missing words in order to check if there is an improvement over the results shown in this paper. Second, as shown in the agreement analysis between annotators, there are some discrepancies in the values assigned for the features, something that potentially affected the results obtained in this paper. Though we proposed using the median value and this yielded robust results for some features, it would be interesting to see other proposals to combine those annotations and mitigate the differences. Finally, the analysis could be enriched if the corpus of sonnets is increased, as well as if the annotations also include sentence-level or stanza-level annotations.

Regarding the usage of the DISCO PAL corpus itself, there are two possible approaches. First, there are research lines that can be pursued related to the psychological tags provided. As we mentioned before, to the best of our knowledge there are no poetry corpora that include annotations regarding psychological states evoked by the poems. This article provides a curated corpus that may help the research regarding the usage of poetry for therapeutic purposes.
This corpus may also help studying the relationship between figurative language (e.g. metaphors) and their contribution to emotions. Even though the presence of figurative language is not explicitly annotated (though it could be included to enhance the corpus), the lexico-semantic features could act as a proxy for it.

The other approach is related to the affective modelling of poetry. DISCO PAL includes 10 affective labels that can be used for studying how to infer the GAM of a Spanish sonnet. This could be accomplished by using Machine Learning models that predict the GAM labels based on the semantic vector of the sonnet.

\begin{acknowledgements}
This work was possible thanks to POSTDATA project: Poetry Standardization and Linked Open Data, Ref.  ERC-2015-STG-679528  project Starting Grant from European Research Council within the horizon H2020. Special thanks to Salvador Ros Mu\~{n}oz, Laura Alises, Marie Olivier and Aroa Rabd\'an.
\end{acknowledgements}

\printbibliography

\onecolumn
\section{Annex}

\begin{table}[h!]
\centering
\resizebox{\textwidth}{!}{%
\begin{tabular}{@{}llllllll@{}}
\toprule
\textbf{variable}    & \textbf{k all} & \textbf{k 12} & \textbf{k 13} & \textbf{k 23} & \textbf{k 1m} & \textbf{k 2m} & \textbf{k 3m} \\ \midrule
Anxiety              & 0.49          & 0.72 & 0.3           & 0.36           & 0.85 & 0.85 & 0.48          \\
Aversion             & 0.57          & 0.72 & 0.5           & 0.47           & 0.89 & 0.82 & 0.64          \\
Depression           & 0.61          & 0.69 & 0.53          & 0.57           & 0.82 & 0.85 & 0.71          \\
Disappointment       & 0.52          & 0.69 & 0.39          & 0.5            & 0.8  & 0.89 & 0.6           \\
Dramatisation        & 0.33          & 0.49 & 0.22          & 0.27           & 0.72 & 0.75 & 0.5           \\
Illusion             & 0.6           & 0.79 & 0.41          & 0.55           & 0.84 & 0.95 & 0.6           \\
Helplessness          & 0.5           & 0.66 & 0.37          & 0.47           & 0.77 & 0.87 & 0.58          \\
Instability          & 0.43          & 0.65 & 0.22          & 0.33           & 0.79 & 0.85 & 0.46          \\
Insecurity           & 0.49          & 0.6  & 0.39          & 0.46           & 0.79 & 0.79 & 0.64          \\
Anger                & 0.57          & 0.82 & 0.41          & 0.44           & 0.92 & 0.89 & 0.53          \\
Obsession            & 0.42          & 0.75 & \textbf{0.13} & 0.23           & 0.85 & 0.89 & 0.29          \\
Pride                & 0.62          & 0.76 & 0.51          & 0.58           & 0.85 & 0.89 & 0.68          \\
Prejudice            & 0.55          & 0.69 & 0.41          & 0.53           & 0.83 & 0.85 & 0.64          \\
Fear (binary)        & 0.51          & 0.66 & 0.39          & 0.45           & 0.81 & 0.84 & 0.6           \\
Vulnerability        & 0.49          & 0.65 & 0.34          & 0.45           & 0.78 & 0.87 & 0.58          \\
concreteness         & 0.26          & 0.55 & \textbf{0.06} & \textbf{0.15}  & 0.75 & 0.78 & 0.27          \\
context availability & 0.25          & 0.64 & \textbf{0.09} & \textbf{0.02}  & 0.88 & 0.76 & \textbf{0.17} \\
Compulsion           & 0.44          & 0.63 & 0.35          & 0.3            & 0.89 & 0.72 & 0.52          \\
Daydream             & 0.44          & 0.55 & 0.29          & 0.45           & 0.66 & 0.86 & 0.58          \\
Grandeur             & 0.53          & 0.66 & 0.35          & 0.56           & 0.72 & 0.94 & 0.62          \\
Idealization         & 0.48          & 0.58 & 0.39          & 0.45           & 0.78 & 0.79 & 0.64          \\
Irritability         & 0.5           & 0.69 & 0.4           & 0.37           & 0.87 & 0.79 & 0.53          \\
Solitude             & 0.58          & 0.76 & 0.44          & 0.51           & 0.83 & 0.92 & 0.59          \\
anger                & 0.38          & 0.6  & 0.27          & 0.26           & 0.77 & 0.8  & 0.45          \\
arousal              & 0.21          & 0.37 & \textbf{0.12} & \textbf{0.11}  & 0.66 & 0.64 & 0.37          \\
disgust              & 0.4           & 0.61 & 0.28          & 0.28           & 0.77 & 0.81 & 0.45          \\
Fear (ordinal)       & 0.34          & 0.53 & 0.22          & 0.28           & 0.67 & 0.8  & 0.47          \\
happiness            & \textbf{0.11} & 0.33 & \textbf{0.05} & \textbf{-0.06} & 0.77 & 0.56 & \textbf{0.2}  \\
imageability         & 0.26          & 0.62 & \textbf{0.09} & \textbf{0.06}  & 0.85 & 0.77 & \textbf{0.2}  \\
sadness              & 0.26          & 0.43 & \textbf{0.19} & \textbf{0.16}  & 0.7  & 0.7  & 0.38          \\
valence              & 0.26          & 0.74 & \textbf{0.02} & \textbf{0.02}  & 0.82 & 0.88 & \textbf{0.11} \\ \bottomrule
\end{tabular}%
}
\caption{K-alpha values between the median value of all the authors compared to each one of them}
\label{table:k-alpha}
\end{table}

\begin{table}[h!]
\centering
\scalebox{0.9}{%
\begin{tabular}{@{}lllllllllllll@{}}

  categories &
  \rot{all lexicons stem} &
  \rot{all lexicons lem} &
  \rot{\cite{perez2021emopro} stem} &
  \rot{\cite{perez2021emopro} lem} &
  \rot{\cite{stadthagen2018norms} stem} &
  \rot{\cite{stadthagen2018norms} lem} &
  \rot{\cite{hinojosa2016affective} stem} &
  \rot{\cite{hinojosa2016affective} lem} &
  \rot{\cite{guasch2016spanish} stem} &
  \rot{\cite{guasch2016spanish} lem} &
  \rot{\cite{ferre2017moved} stem} &
  \rot{\cite{ferre2017moved} lem} \\ \midrule
all            & 0.68 & 0.56 & 0.09 & 0.08 & 0.51 & 0.35 & 0.11 & 0.06 & 0.14 & 0.09 & 0.23 & 0.16 \\
Anxiety        & 0.75 & 0.65 & 0.12 & 0.1  & 0.56 & 0.4  & 0.13 & 0.07 & 0.17 & 0.12 & 0.28 & 0.21 \\
Aversion       & 0.73 & 0.64 & 0.11 & 0.09 & 0.55 & 0.39 & 0.13 & 0.08 & 0.16 & 0.11 & 0.25 & 0.19 \\
Depression     & 0.79 & 0.68 & 0.15 & 0.12 & 0.58 & 0.4  & 0.14 & 0.09 & 0.18 & 0.12 & 0.3  & 0.22 \\
Disappointment & 0.78 & 0.68 & 0.13 & 0.1  & 0.57 & 0.4  & 0.15 & 0.08 & 0.19 & 0.13 & 0.3  & 0.22 \\
Dramatisation  & 0.72 & 0.62 & 0.11 & 0.1  & 0.55 & 0.37 & 0.13 & 0.08 & 0.16 & 0.11 & 0.25 & 0.19 \\
Illusion       & 0.75 & 0.67 & 0.11 & 0.09 & 0.56 & 0.41 & 0.13 & 0.08 & 0.18 & 0.12 & 0.28 & 0.2  \\
Helplessness    & 0.75 & 0.65 & 0.13 & 0.11 & 0.56 & 0.39 & 0.14 & 0.08 & 0.17 & 0.12 & 0.28 & 0.21 \\
Instability    & 0.77 & 0.66 & 0.13 & 0.11 & 0.57 & 0.39 & 0.13 & 0.08 & 0.18 & 0.12 & 0.29 & 0.21 \\
Insecurity     & 0.77 & 0.69 & 0.14 & 0.12 & 0.57 & 0.41 & 0.14 & 0.08 & 0.18 & 0.12 & 0.29 & 0.22 \\
Anger          & 0.76 & 0.67 & 0.13 & 0.11 & 0.56 & 0.4  & 0.13 & 0.08 & 0.18 & 0.13 & 0.29 & 0.22 \\
Obsession      & 0.76 & 0.67 & 0.14 & 0.11 & 0.56 & 0.4  & 0.14 & 0.09 & 0.18 & 0.12 & 0.29 & 0.22 \\
Pride          & 0.76 & 0.67 & 0.11 & 0.09 & 0.57 & 0.41 & 0.13 & 0.09 & 0.18 & 0.12 & 0.28 & 0.2  \\
Prejudice      & 0.78 & 0.7  & 0.11 & 0.09 & 0.57 & 0.41 & 0.15 & 0.09 & 0.19 & 0.13 & 0.31 & 0.24 \\
Fear (binary)  & 0.75 & 0.64 & 0.12 & 0.1  & 0.56 & 0.39 & 0.13 & 0.08 & 0.16 & 0.11 & 0.27 & 0.19 \\
Vulnerability  & 0.73 & 0.61 & 0.12 & 0.09 & 0.54 & 0.37 & 0.13 & 0.07 & 0.16 & 0.1  & 0.26 & 0.18 \\
Compulsion     & 0.76 & 0.66 & 0.13 & 0.1  & 0.56 & 0.4  & 0.13 & 0.08 & 0.18 & 0.12 & 0.28 & 0.2  \\
Daydream       & 0.77 & 0.67 & 0.11 & 0.09 & 0.57 & 0.4  & 0.14 & 0.09 & 0.17 & 0.13 & 0.27 & 0.21 \\
Grandeur       & 0.72 & 0.62 & 0.11 & 0.09 & 0.54 & 0.38 & 0.12 & 0.08 & 0.16 & 0.11 & 0.26 & 0.19 \\
Idealization   & 0.73 & 0.63 & 0.11 & 0.09 & 0.54 & 0.39 & 0.13 & 0.08 & 0.16 & 0.11 & 0.26 & 0.19 \\
Irritability   & 0.79 & 0.69 & 0.14 & 0.11 & 0.59 & 0.42 & 0.13 & 0.07 & 0.19 & 0.13 & 0.3  & 0.23 \\
Solitude       & 0.76 & 0.66 & 0.13 & 0.11 & 0.56 & 0.39 & 0.14 & 0.08 & 0.17 & 0.12 & 0.28 & 0.21 \\ \bottomrule
\end{tabular}%
}
\caption{Fraction of words (after lemmatization or stemming) from the original DISCO PAL corpus (total and per psychological category) in the different source lexicons}
\label{table:fraction-words}
\end{table}

\begin{table}[h!]
\centering
\resizebox{\textwidth}{!}{%
\begin{tabular}{@{}lllllllllllll@{}}
\toprule
\textbf{Category} &
  \textbf{Feature} &
  \textbf{Feature (GAM)} &
  \textbf{r2} &
  \textbf{coeff} &
  \textbf{sign} &
   &
  \textbf{Category} &
  \textbf{Feature} &
  \textbf{Feature (GAM)} &
  \textbf{r2} &
  \textbf{coeff} &
  \textbf{sign} \\ \midrule
Anxiety        & valence              & valence\_mean      & 0.89 & 1.97 & yes &  & Prejudice     & valence              & valence\_mean      & 0.94 & 1.89 & yes \\
Anxiety        & arousal              & arousal\_mean      & 0.92 & 2.02 & yes &  & Prejudice     & arousal              & arousal\_mean      & 0.89 & 2.11 & yes \\
Anxiety        & happiness            & happiness\_mean    & 0.88 & 1.67 & yes &  & Prejudice     & happiness            & happiness\_mean    & 0.84 & 1.52 & yes \\
Anxiety        & anger                & anger\_mean        & 0.77 & 0.73 & yes &  & Prejudice     & anger                & anger\_mean        & 0.82 & 0.67 & yes \\
Anxiety        & sadness              & sadness\_mean      & 0.9  & 0.62 & yes &  & Prejudice     & sadness              & sadness\_mean      & 0.87 & 0.72 & yes \\
Anxiety        & fear                 & fear\_mean         & 0.84 & 0.91 & yes &  & Prejudice     & fear                 & fear\_mean         & 0.8  & 0.99 & yes \\
Anxiety        & disgust              & disgust\_mean      & 0.84 & 0.62 & yes &  & Prejudice     & disgust              & disgust\_mean      & 0.87 & 0.59 & yes \\
Anxiety        & concreteness         & concreteness\_mean & 0.8  & 1.79 & yes &  & Prejudice     & concreteness         & concreteness\_mean & 0.74 & 1.97 & yes \\
Anxiety        & imageability         & imageability\_mean & 0.76 & 1.98 & yes &  & Prejudice     & imageability         & imageability\_mean & 0.72 & 1.94 & yes \\
Anxiety        & context availability & cont\_ava\_mean    & 0.77 & 2.19 & yes &  & Prejudice     & context availability & cont\_ava\_mean    & 0.74 & 2.2  & yes \\
Aversion       & valence              & valence\_mean      & 0.92 & 1.99 & yes &  & Fear (binary) & valence              & valence\_mean      & 0.92 & 1.89 & yes \\
Aversion       & arousal              & arousal\_mean      & 0.92 & 2.02 & yes &  & Fear (binary) & arousal              & arousal\_mean      & 0.91 & 2.08 & yes \\
Aversion       & happiness            & happiness\_mean    & 0.88 & 1.78 & yes &  & Fear (binary) & happiness            & happiness\_mean    & 0.86 & 1.66 & yes \\
Aversion       & anger                & anger\_mean        & 0.78 & 0.74 & yes &  & Fear (binary) & anger                & anger\_mean        & 0.78 & 0.88 & yes \\
Aversion       & sadness              & sadness\_mean      & 0.9  & 0.65 & yes &  & Fear (binary) & sadness              & sadness\_mean      & 0.9  & 0.64 & yes \\
Aversion       & fear                 & fear\_mean         & 0.81 & 1.03 & yes &  & Fear (binary) & fear                 & fear\_mean         & 0.82 & 0.94 & yes \\
Aversion       & disgust              & disgust\_mean      & 0.88 & 0.6  & yes &  & Fear (binary) & disgust              & disgust\_mean      & 0.83 & 0.63 & yes \\
Aversion       & concreteness         & concreteness\_mean & 0.78 & 1.89 & yes &  & Fear (binary) & concreteness         & concreteness\_mean & 0.79 & 1.76 & yes \\
Aversion       & imageability         & imageability\_mean & 0.75 & 1.98 & yes &  & Fear (binary) & imageability         & imageability\_mean & 0.8  & 1.97 & yes \\
Aversion       & context availability & cont\_ava\_mean    & 0.76 & 2.28 & yes &  & Fear (binary) & context availability & cont\_ava\_mean    & 0.78 & 2.23 & yes \\
Depression     & valence              & valence\_mean      & 0.88 & 2.01 & yes &  & Vulnerability & valence              & valence\_mean      & 0.91 & 1.84 & yes \\
Depression     & arousal              & arousal\_mean      & 0.95 & 2.06 & yes &  & Vulnerability & arousal              & arousal\_mean      & 0.92 & 2.08 & yes \\
Depression     & happiness            & happiness\_mean    & 0.85 & 1.6  & yes &  & Vulnerability & happiness            & happiness\_mean    & 0.86 & 1.61 & yes \\
Depression     & anger                & anger\_mean        & 0.82 & 0.87 & yes &  & Vulnerability & anger                & anger\_mean        & 0.78 & 0.87 & yes \\
Depression     & sadness              & sadness\_mean      & 0.94 & 0.58 & yes &  & Vulnerability & sadness              & sadness\_mean      & 0.91 & 0.64 & yes \\
Depression     & fear                 & fear\_mean         & 0.88 & 1.07 & yes &  & Vulnerability & fear                 & fear\_mean         & 0.83 & 0.99 & yes \\
Depression     & disgust              & disgust\_mean      & 0.87 & 0.63 & yes &  & Vulnerability & disgust              & disgust\_mean      & 0.84 & 0.66 & yes \\
Depression     & concreteness         & concreteness\_mean & 0.81 & 2.16 & yes &  & Vulnerability & concreteness         & concreteness\_mean & 0.78 & 1.76 & yes \\
Depression     & imageability         & imageability\_mean & 0.86 & 2.38 & yes &  & Vulnerability & imageability         & imageability\_mean & 0.77 & 1.95 & yes \\
Depression     & context availability & cont\_ava\_mean    & 0.86 & 2.78 & yes &  & Vulnerability & context availability & cont\_ava\_mean    & 0.77 & 2.19 & yes \\
Disappointment & valence              & valence\_mean      & 0.91 & 1.92 & yes &  & Compulsion    & valence              & valence\_mean      & 0.94 & 1.68 & yes \\
Disappointment & arousal              & arousal\_mean      & 0.94 & 2.05 & yes &  & Compulsion    & arousal              & arousal\_mean      & 0.92 & 2.23 & yes \\
Disappointment & happiness            & happiness\_mean    & 0.85 & 1.63 & yes &  & Compulsion    & happiness            & happiness\_mean    & 0.83 & 1.3  & yes \\
Disappointment & anger                & anger\_mean        & 0.8  & 0.73 & yes &  & Compulsion    & anger                & anger\_mean        & 0.82 & 0.92 & yes \\
Disappointment & sadness              & sadness\_mean      & 0.92 & 0.6  & yes &  & Compulsion    & sadness              & sadness\_mean      & 0.86 & 0.73 & yes \\
Disappointment & fear                 & fear\_mean         & 0.86 & 1.06 & yes &  & Compulsion    & fear                 & fear\_mean         & 0.84 & 1.19 & yes \\
Disappointment & disgust              & disgust\_mean      & 0.89 & 0.6  & yes &  & Compulsion    & disgust              & disgust\_mean      & 0.81 & 0.71 & yes \\
Disappointment & concreteness         & concreteness\_mean & 0.81 & 2.26 & yes &  & Compulsion    & concreteness         & concreteness\_mean & 0.78 & 1.94 & yes \\
Disappointment & imageability         & imageability\_mean & 0.78 & 2.31 & yes &  & Compulsion    & imageability         & imageability\_mean & 0.79 & 2.11 & yes \\
Disappointment & context availability & cont\_ava\_mean    & 0.77 & 2.58 & yes &  & Compulsion    & context availability & cont\_ava\_mean    & 0.8  & 2.47 & yes \\
Dramatisation  & valence              & valence\_mean      & 0.92 & 1.76 & yes &  & Daydream      & valence              & valence\_mean      & 0.95 & 1.6  & yes \\
Dramatisation  & arousal              & arousal\_mean      & 0.89 & 2.17 & yes &  & Daydream      & arousal              & arousal\_mean      & 0.9  & 2.29 & yes \\
Dramatisation  & happiness            & happiness\_mean    & 0.87 & 1.73 & yes &  & Daydream      & happiness            & happiness\_mean    & 0.82 & 1.15 & yes \\
Dramatisation  & anger                & anger\_mean        & 0.79 & 0.94 & yes &  & Daydream      & anger                & anger\_mean        & 0.86 & 1.11 & yes \\
Dramatisation  & sadness              & sadness\_mean      & 0.86 & 0.68 & yes &  & Daydream      & sadness              & sadness\_mean      & 0.8  & 0.77 & yes \\
Dramatisation  & fear                 & fear\_mean         & 0.82 & 1.07 & yes &  & Daydream      & fear                 & fear\_mean         & 0.83 & 1.11 & yes \\
Dramatisation  & disgust              & disgust\_mean      & 0.84 & 0.7  & yes &  & Daydream      & disgust              & disgust\_mean      & 0.8  & 0.89 & yes \\
Dramatisation  & concreteness         & concreteness\_mean & 0.8  & 1.76 & yes &  & Daydream      & concreteness         & concreteness\_mean & 0.85 & 1.53 & yes \\
Dramatisation  & imageability         & imageability\_mean & 0.78 & 1.93 & yes &  & Daydream      & imageability         & imageability\_mean & 0.83 & 1.69 & yes \\
Dramatisation  & context availability & cont\_ava\_mean    & 0.78 & 2.17 & yes &  & Daydream      & context availability & cont\_ava\_mean    & 0.83 & 1.84 & yes \\
Illusion       & valence              & valence\_mean      & 0.96 & 1.56 & yes &  & Grandeur      & valence              & valence\_mean      & 0.95 & 1.56 & yes \\
Illusion       & arousal              & arousal\_mean      & 0.88 & 2.27 & yes &  & Grandeur      & arousal              & arousal\_mean      & 0.87 & 2.26 & yes \\
Illusion       & happiness            & happiness\_mean    & 0.82 & 1.02 & yes &  & Grandeur      & happiness            & happiness\_mean    & 0.8  & 1.16 & yes \\
Illusion       & anger                & anger\_mean        & 0.8  & 0.98 & yes &  & Grandeur      & anger                & anger\_mean        & 0.84 & 1.08 & yes \\
Illusion       & sadness              & sadness\_mean      & 0.8  & 0.78 & yes &  & Grandeur      & sadness              & sadness\_mean      & 0.81 & 0.82 & yes \\
Illusion       & fear                 & fear\_mean         & 0.85 & 1.15 & yes &  & Grandeur      & fear                 & fear\_mean         & 0.85 & 1.17 & yes \\
Illusion       & disgust              & disgust\_mean      & 0.8  & 0.87 & yes &  & Grandeur      & disgust              & disgust\_mean      & 0.8  & 0.87 & yes \\
Illusion       & concreteness         & concreteness\_mean & 0.81 & 1.69 & yes &  & Grandeur      & concreteness         & concreteness\_mean & 0.8  & 1.64 & yes \\
Illusion       & imageability         & imageability\_mean & 0.78 & 1.81 & yes &  & Grandeur      & imageability         & imageability\_mean & 0.78 & 1.78 & yes \\
Illusion       & context availability & cont\_ava\_mean    & 0.81 & 2.07 & yes &  & Grandeur      & context availability & cont\_ava\_mean    & 0.8  & 2.01 & yes \\
Helplessness    & valence              & valence\_mean      & 0.89 & 1.98 & yes &  & Idealization  & valence              & valence\_mean      & 0.95 & 1.61 & yes \\
Helplessness    & arousal              & arousal\_mean      & 0.93 & 1.99 & yes &  & Idealization  & arousal              & arousal\_mean      & 0.88 & 2.22 & yes \\
Helplessness    & happiness            & happiness\_mean    & 0.91 & 1.83 & yes &  & Idealization  & happiness            & happiness\_mean    & 0.81 & 1.25 & yes \\
Helplessness    & anger                & anger\_mean        & 0.77 & 0.78 & yes &  & Idealization  & anger                & anger\_mean        & 0.82 & 1.01 & yes \\
Helplessness    & sadness              & sadness\_mean      & 0.91 & 0.61 & yes &  & Idealization  & sadness              & sadness\_mean      & 0.82 & 0.75 & yes \\
Helplessness    & fear                 & fear\_mean         & 0.83 & 0.88 & yes &  & Idealization  & fear                 & fear\_mean         & 0.86 & 1.15 & yes \\
Helplessness    & disgust              & disgust\_mean      & 0.87 & 0.59 & yes &  & Idealization  & disgust              & disgust\_mean      & 0.79 & 0.8  & yes \\
Helplessness    & concreteness         & concreteness\_mean & 0.77 & 1.97 & yes &  & Idealization  & concreteness         & concreteness\_mean & 0.78 & 1.68 & yes \\
Helplessness    & imageability         & imageability\_mean & 0.81 & 2.4  & yes &  & Idealization  & imageability         & imageability\_mean & 0.78 & 1.82 & yes \\
Helplessness    & context availability & cont\_ava\_mean    & 0.79 & 2.62 & yes &  & Idealization  & context availability & cont\_ava\_mean    & 0.78 & 2.03 & yes \\
Instability    & valence              & valence\_mean      & 0.9  & 1.89 & yes &  & Irritability  & valence              & valence\_mean      & 0.93 & 2.13 & yes \\
Instability    & arousal              & arousal\_mean      & 0.91 & 1.99 & yes &  & Irritability  & arousal              & arousal\_mean      & 0.92 & 2    & yes \\
Instability    & happiness            & happiness\_mean    & 0.95 & 1.95 & yes &  & Irritability  & happiness            & happiness\_mean    & 0.97 & 2.11 & yes \\
Instability    & anger                & anger\_mean        & 0.78 & 0.83 & yes &  & Irritability  & anger                & anger\_mean        & 0.87 & 0.64 & yes \\
Instability    & sadness              & sadness\_mean      & 0.91 & 0.63 & yes &  & Irritability  & sadness              & sadness\_mean      & 0.92 & 0.64 & yes \\
Instability    & fear                 & fear\_mean         & 0.84 & 1.03 & yes &  & Irritability  & fear                 & fear\_mean         & 0.82 & 0.96 & yes \\
Instability    & disgust              & disgust\_mean      & 0.87 & 0.67 & yes &  & Irritability  & disgust              & disgust\_mean      & 0.92 & 0.55 & yes \\
Instability    & concreteness         & concreteness\_mean & 0.8  & 1.71 & yes &  & Irritability  & concreteness         & concreteness\_mean & 0.73 & 1.84 & yes \\
Instability    & imageability         & imageability\_mean & 0.76 & 1.79 & yes &  & Irritability  & imageability         & imageability\_mean & 0.69 & 1.75 & yes \\
Instability    & context availability & cont\_ava\_mean    & 0.75 & 2.01 & yes &  & Irritability  & context availability & cont\_ava\_mean    & 0.7  & 1.93 & yes \\
Insecurity     & valence              & valence\_mean      & 0.9  & 1.79 & yes &  & Solitude      & valence              & valence\_mean      & 0.89 & 1.87 & yes \\
Insecurity     & arousal              & arousal\_mean      & 0.92 & 2.06 & yes &  & Solitude      & arousal              & arousal\_mean      & 0.89 & 1.9  & yes \\
Insecurity     & happiness            & happiness\_mean    & 0.91 & 1.72 & yes &  & Solitude      & happiness            & happiness\_mean    & 0.77 & 1.35 & yes \\
Insecurity     & anger                & anger\_mean        & 0.79 & 0.95 & yes &  & Solitude      & anger                & anger\_mean        & 0.8  & 0.84 & yes \\
Insecurity     & sadness              & sadness\_mean      & 0.88 & 0.67 & yes &  & Solitude      & sadness              & sadness\_mean      & 0.92 & 0.59 & yes \\
Insecurity     & fear                 & fear\_mean         & 0.82 & 0.92 & yes &  & Solitude      & fear                 & fear\_mean         & 0.88 & 1.01 & yes \\
Insecurity     & disgust              & disgust\_mean      & 0.87 & 0.72 & yes &  & Solitude      & disgust              & disgust\_mean      & 0.83 & 0.67 & yes \\
Insecurity     & concreteness         & concreteness\_mean & 0.77 & 1.8  & yes &  & Solitude      & concreteness         & concreteness\_mean & 0.83 & 1.85 & yes \\
Insecurity     & imageability         & imageability\_mean & 0.74 & 1.92 & yes &  & Solitude      & imageability         & imageability\_mean & 0.79 & 2.04 & yes \\
Insecurity     & context availability & cont\_ava\_mean    & 0.75 & 2.23 & yes &  & Solitude      & context availability & cont\_ava\_mean    & 0.8  & 2.33 & yes \\
Pride          & valence              & valence\_mean      & 0.95 & 1.58 & yes &  &               &                      &                    &      &      &     \\
Pride          & arousal              & arousal\_mean      & 0.87 & 2.26 & yes &  &               &                      &                    &      &      &     \\
Pride          & happiness            & happiness\_mean    & 0.82 & 1.26 & yes &  &               &                      &                    &      &      &     \\
Pride          & anger                & anger\_mean        & 0.76 & 0.86 & yes &  &               &                      &                    &      &      &     \\
Pride          & sadness              & sadness\_mean      & 0.81 & 0.82 & yes &  &               &                      &                    &      &      &     \\
Pride          & fear                 & fear\_mean         & 0.9  & 1.35 & yes &  &               &                      &                    &      &      &     \\
Pride          & disgust              & disgust\_mean      & 0.76 & 0.77 & yes &  &               &                      &                    &      &      &     \\
Pride          & concreteness         & concreteness\_mean & 0.79 & 1.64 & yes &  &               &                      &                    &      &      &     \\
Pride          & imageability         & imageability\_mean & 0.77 & 1.75 & yes &  &               &                      &                    &      &      &     \\
Pride          & context availability & cont\_ava\_mean    & 0.78 & 1.94 & yes &  &               &                      &                    &      &      &     \\ \bottomrule
\end{tabular}%
}
\caption{Partial dependence analysis between annotated features and their inferred GAM counterpart (for each psychological category)}
\label{table:partial-analysis-per-psycho-tag}
\end{table}

\begin{table}[h!]
\centering
\resizebox{\textwidth}{!}{%
\begin{tabular}{@{}lllllllllll@{}}
\toprule
\textbf{Category} &
  \textbf{Feature (GAM)} &
  \textbf{M (=1)} &
  \textbf{M (=0)} &
  \textbf{p} &
   &
  \textbf{Category} &
  \textbf{Feature (GAM)} &
  \textbf{M (=1)} &
  \textbf{M (=0)} &
  \textbf{p} \\ \midrule
Solitude     & valence\_mean      & 5.23 & 5.34 & 0.0485 &  & Depression     & anger\_mean        & 1.71 & 1.62 & 0.0015 \\
Solitude     & sadness\_mean      & 1.87 & 1.74 & 0      &  & Depression     & sadness\_mean      & 1.94 & 1.74 & 0      \\
Solitude     & fear\_mean         & 1.8  & 1.75 & 0.0394 &  & Depression     & fear\_mean         & 1.85 & 1.74 & 0.0001 \\
Illusion     & valence\_mean      & 5.55 & 5.22 & 0      &  & Disappointment & valence\_mean      & 5.2  & 5.33 & 0.0234 \\
Illusion     & arousal\_mean      & 5.18 & 5.3  & 0      &  & Disappointment & arousal\_mean      & 5.33 & 5.26 & 0.0386 \\
Illusion     & happiness\_mean    & 2.48 & 2.29 & 0      &  & Disappointment & anger\_mean        & 1.69 & 1.63 & 0.0236 \\
Illusion     & anger\_mean        & 1.55 & 1.67 & 0      &  & Disappointment & sadness\_mean      & 1.87 & 1.75 & 0.0001 \\
Illusion     & sadness\_mean      & 1.67 & 1.81 & 0      &  & Disappointment & fear\_mean         & 1.82 & 1.75 & 0.0057 \\
Illusion     & fear\_mean         & 1.69 & 1.78 & 0      &  & Aversion       & valence\_mean      & 5.16 & 5.39 & 0      \\
Illusion     & disgust\_mean      & 1.38 & 1.47 & 0      &  & Aversion       & arousal\_mean      & 5.31 & 5.24 & 0.0113 \\
Daydream     & valence\_mean      & 5.58 & 5.26 & 0      &  & Aversion       & happiness\_mean    & 2.24 & 2.39 & 0      \\
Daydream     & arousal\_mean      & 5.18 & 5.28 & 0.005  &  & Aversion       & anger\_mean        & 1.71 & 1.59 & 0      \\
Daydream     & happiness\_mean    & 2.48 & 2.31 & 0      &  & Aversion       & sadness\_mean      & 1.83 & 1.74 & 0.0003 \\
Daydream     & anger\_mean        & 1.52 & 1.66 & 0      &  & Aversion       & fear\_mean         & 1.79 & 1.74 & 0.0358 \\
Daydream     & sadness\_mean      & 1.64 & 1.8  & 0      &  & Aversion       & disgust\_mean      & 1.48 & 1.42 & 0.0005 \\
Daydream     & fear\_mean         & 1.68 & 1.77 & 0.0003 &  & Aversion       & imageability\_mean & 4.24 & 4.42 & 0.0176 \\
Daydream     & disgust\_mean      & 1.36 & 1.46 & 0      &  & Insecurity     & valence\_mean      & 5.1  & 5.35 & 0.0001 \\
Grandeur     & valence\_mean      & 5.5  & 5.19 & 0      &  & Insecurity     & arousal\_mean      & 5.35 & 5.25 & 0.0041 \\
Grandeur     & arousal\_mean      & 5.23 & 5.29 & 0.0121 &  & Insecurity     & happiness\_mean    & 2.22 & 2.36 & 0.0002 \\
Grandeur     & happiness\_mean    & 2.43 & 2.28 & 0      &  & Insecurity     & anger\_mean        & 1.74 & 1.62 & 0      \\
Grandeur     & anger\_mean        & 1.57 & 1.68 & 0      &  & Insecurity     & sadness\_mean      & 1.89 & 1.75 & 0      \\
Grandeur     & sadness\_mean      & 1.67 & 1.83 & 0      &  & Insecurity     & fear\_mean         & 1.86 & 1.74 & 0      \\
Grandeur     & fear\_mean         & 1.7  & 1.79 & 0      &  & Insecurity     & disgust\_mean      & 1.49 & 1.44 & 0.0097 \\
Grandeur     & disgust\_mean      & 1.42 & 1.46 & 0.0033 &  & Helplessness    & valence\_mean      & 5.17 & 5.35 & 0.0006 \\
Pride        & valence\_mean      & 5.51 & 5.24 & 0      &  & Helplessness    & arousal\_mean      & 5.35 & 5.24 & 0.0009 \\
Pride        & arousal\_mean      & 5.22 & 5.29 & 0.0273 &  & Helplessness    & happiness\_mean    & 2.27 & 2.36 & 0.0064 \\
Pride        & happiness\_mean    & 2.42 & 2.3  & 0.0001 &  & Helplessness    & anger\_mean        & 1.72 & 1.61 & 0      \\
Pride        & anger\_mean        & 1.57 & 1.66 & 0      &  & Helplessness    & sadness\_mean      & 1.87 & 1.74 & 0      \\
Pride        & sadness\_mean      & 1.67 & 1.81 & 0      &  & Helplessness    & fear\_mean         & 1.83 & 1.74 & 0.0001 \\
Pride        & fear\_mean         & 1.68 & 1.78 & 0      &  & Vulnerability  & valence\_mean      & 5.18 & 5.42 & 0      \\
Pride        & disgust\_mean      & 1.42 & 1.46 & 0.0325 &  & Vulnerability  & arousal\_mean      & 5.31 & 5.23 & 0.0008 \\
Pride        & concreteness\_mean & 4.28 & 4.36 & 0.0452 &  & Vulnerability  & happiness\_mean    & 2.28 & 2.39 & 0      \\
Irritability & valence\_mean      & 5.09 & 5.34 & 0.0002 &  & Vulnerability  & anger\_mean        & 1.7  & 1.58 & 0      \\
Irritability & arousal\_mean      & 5.34 & 5.26 & 0.0437 &  & Vulnerability  & sadness\_mean      & 1.85 & 1.7  & 0      \\
Irritability & happiness\_mean    & 2.22 & 2.35 & 0.0005 &  & Vulnerability  & fear\_mean         & 1.82 & 1.7  & 0      \\
Irritability & anger\_mean        & 1.73 & 1.62 & 0.0002 &  & Vulnerability  & disgust\_mean      & 1.47 & 1.43 & 0.0088 \\
Irritability &
  sadness\_mean &
  1.89 &
  1.75 &
  0.0001 &
   &
  Vulnerability &
  imageability\_mean &
  4.24 &
  4.46 &
  0.0025 \\
Anxiety      & valence\_mean      & 5.18 & 5.36 & 0.0005 &  & Fear (binary)  & valence\_mean      & 5.18 & 5.38 & 0      \\
Anxiety      & arousal\_mean      & 5.34 & 5.24 & 0.0008 &  & Fear (binary)  & arousal\_mean      & 5.32 & 5.24 & 0.0038 \\
Anxiety      & happiness\_mean    & 2.27 & 2.36 & 0.0035 &  & Fear (binary)  & happiness\_mean    & 2.27 & 2.37 & 0.0002 \\
Anxiety      & anger\_mean        & 1.7  & 1.61 & 0.0001 &  & Fear (binary)  & anger\_mean        & 1.69 & 1.61 & 0      \\
Anxiety      & sadness\_mean      & 1.86 & 1.73 & 0      &  & Fear (binary)  & sadness\_mean      & 1.84 & 1.73 & 0      \\
Anxiety      & fear\_mean         & 1.83 & 1.73 & 0      &  & Fear (binary)  & fear\_mean         & 1.81 & 1.73 & 0.0003 \\
Anger        & valence\_mean      & 5.11 & 5.36 & 0      &  & Fear (binary)  & disgust\_mean      & 1.47 & 1.43 & 0.0271 \\
Anger        & arousal\_mean      & 5.34 & 5.25 & 0.0046 &  & Obsession      & valence\_mean      & 5.1  & 5.34 & 0.0009 \\
Anger        & happiness\_mean    & 2.24 & 2.36 & 0.0001 &  & Obsession      & arousal\_mean      & 5.35 & 5.26 & 0.0184 \\
Anger        & anger\_mean        & 1.73 & 1.61 & 0      &  & Obsession      & happiness\_mean    & 2.25 & 2.35 & 0.0162 \\
Anger        & sadness\_mean      & 1.87 & 1.74 & 0      &  & Obsession      & anger\_mean        & 1.74 & 1.62 & 0.0002 \\
Anger        & fear\_mean         & 1.82 & 1.74 & 0.0012 &  & Obsession      & sadness\_mean      & 1.9  & 1.75 & 0.0001 \\
Anger        & disgust\_mean      & 1.49 & 1.43 & 0.0093 &  & Obsession      & fear\_mean         & 1.82 & 1.75 & 0.0277 \\
Instability  & valence\_mean      & 5.08 & 5.38 & 0      &  & Obsession      & imageability\_mean & 4.14 & 4.38 & 0.0355 \\
Instability  & arousal\_mean      & 5.37 & 5.24 & 0      &  & Compulsion     & imageability\_mean & 4.21 & 4.4  & 0.0379 \\
Instability  & happiness\_mean    & 2.22 & 2.37 & 0      &  & Prejudice      & disgust\_mean      & 1.5  & 1.44 & 0.0172 \\
Instability  & anger\_mean        & 1.75 & 1.6  & 0      &  & Dramatisation  & valence\_mean      & 5.22 & 5.37 & 0.0018 \\
Instability  & sadness\_mean      & 1.9  & 1.73 & 0      &  & Dramatisation  & arousal\_mean      & 5.31 & 5.24 & 0.0114 \\
Instability  & fear\_mean         & 1.85 & 1.73 & 0      &  & Dramatisation  & happiness\_mean    & 2.29 & 2.37 & 0.003  \\
Instability  & disgust\_mean      & 1.5  & 1.43 & 0.0003 &  & Dramatisation  & anger\_mean        & 1.69 & 1.6  & 0      \\
Idealization & valence\_mean      & 5.49 & 5.19 & 0      &  & Dramatisation  & sadness\_mean      & 1.82 & 1.74 & 0.0009 \\
Idealization & arousal\_mean      & 5.21 & 5.3  & 0.0011 &  & Dramatisation  & fear\_mean         & 1.8  & 1.73 & 0.001  \\
Idealization & happiness\_mean    & 2.44 & 2.27 & 0      &  & Dramatisation  & disgust\_mean      & 1.47 & 1.43 & 0.0233 \\
Idealization & anger\_mean        & 1.56 & 1.68 & 0      &  &                &                    &      &      &        \\
Idealization & sadness\_mean      & 1.68 & 1.83 & 0      &  &                &                    &      &      &        \\
Idealization & fear\_mean         & 1.69 & 1.8  & 0      &  &                &                    &      &      &        \\
Idealization & disgust\_mean      & 1.4  & 1.47 & 0      &  &                &                    &      &      &        \\
Depression   & valence\_mean      & 5.12 & 5.34 & 0.0009 &  &                &                    &      &      &        \\
Depression   & arousal\_mean      & 5.36 & 5.25 & 0.0061 &  &                &                    &      &      &        \\
Depression   & happiness\_mean    & 2.27 & 2.35 & 0.0325 &  &                &                    &      &      &        \\ \bottomrule
\end{tabular}%
}
\caption{One-way ANOVA between GAM values according to their psychological category}
\label{table:oneway-anova}
\end{table}




\end{document}